\documentclass[11pt,a4paper]{article}

\usepackage[margin=2.5cm]{geometry}
\usepackage[utf8]{inputenc}
\usepackage[T1]{fontenc}
\usepackage{amssymb}
\usepackage{amsmath}
\usepackage{amsthm}
\usepackage{graphicx}
\usepackage{xspace}
\usepackage{siunitx}
\usepackage[export]{adjustbox} 

\usepackage{tikz}
\usetikzlibrary{fit,arrows.meta,positioning,calc,shadows,backgrounds,matrix}
\tikzset{
  seq/.style={draw=black!60, line width=0.8pt, rounded corners=3pt, fill=white,
              inner sep=5pt, minimum height=8mm, align=left,
              drop shadow={opacity=0.12, shadow xshift=0.6pt, shadow yshift=-0.9pt}},
  res/.style={draw=blue!45!black, line width=0.8pt, rounded corners=3pt, fill=blue!9,
              inner sep=5pt, minimum height=8mm, align=left},
  stage/.style={draw=#1, line width=0.9pt, rounded corners=3pt, fill=white,
              minimum height=7.5mm, align=center, inner sep=3pt, font=\small,
              drop shadow={opacity=0.12, shadow xshift=0.6pt, shadow yshift=-0.9pt}},
  cstage/.style={stage=#1, dashed, fill=#1!12},
  arr/.style={->, line width=0.9pt, draw=black!75},
  group/.style={draw=black!60, rounded corners=4pt, densely dotted, line width=0.6pt, inner sep=5pt},
  lanelab/.style={font=\scriptsize\itshape, text=black!55, inner sep=2pt},
  cell/.style={draw=black!55, line width=0.5pt, fill=white, minimum size=4.6mm, inner sep=0pt, anchor=west, font=\scriptsize},
  cdel/.style={cell, fill=black!14, text=black!60},
  cnt/.style={cell, fill=green!16},
  cB/.style={cell, fill=red!14, text=red!60!black},
  cdelta/.style={cell, fill=blue!12, text=blue!50!black},
  cellw/.style={cell, minimum width=7.8mm, minimum height=4.6mm},
  cdelw/.style={cellw, fill=black!14, text=black!60},
  cellwhl/.style={cellw, fill=orange!18},
  cdelwhl/.style={cdelw},
  cellwhl2/.style={cellw, fill=orange!18, draw=orange!70!black, line width=0.9pt},
  pill/.style={font=\scriptsize, draw=green!45!black, fill=green!8, rounded corners=6pt, inner xsep=5pt, inner ysep=2pt},
}

\newcommand{\fitwidth}[1]{\resizebox{\ifdim\width>\textwidth\textwidth\else\width\fi}{!}{#1}}
\usepackage{multirow}
\usepackage{comment}
\usepackage{url}
\usepackage{float}
\usepackage{xcolor}
\usepackage{arydshln}
\usepackage{algorithm}
\usepackage{algpseudocode}
\newtheorem{theorem}{Theorem}
\newtheorem{lemma}{Lemma}
\newtheorem{proposition}{Proposition}
\newtheorem{definition}{Definition}
\theoremstyle{remark}

\newcommand{\eg}{{\it e.g.}\xspace}

\newcommand{\etal}{{\it et al.}\xspace}

\newcommand{\Repair}{\textsf{RePair}\xspace}
\newcommand{\mmRepair}{\textsf{mm-RePair}\xspace}
\newcommand{\our}{\textsf{re-hybd}\xspace}
\newcommand{\reans}{\textsf{re-ans}\xspace}
\newcommand{\reansPC}{\textsf{re-pc}\xspace}
\newcommand{\ourPC}{\textsf{re-hybd-pc}\xspace}
\newcommand{\ansfold}{\texttt{ans-fold}\xspace}
\newcommand{\wcode}{\textsf{wcode}\xspace}
\newcommand{\pgsf}{\texttt{pgsf}\xspace}

\usepackage[hidelinks]{hyperref}

\title{Hybrid Compression for Memory-Efficient\\ Matrix--Vector Multiplication%
\thanks{A preliminary version of this work was accepted for presentation at the IEEE Data Compression Conference (DCC 2027)~\cite{LouzaMT2027}.}
}

\author{%
  Felipe A. Louza$^{1,2}$\thanks{Corresponding author.}\\
  {\small\texttt{louza@ufu.br}}
  \and
  Giovanni Manzini$^{3}$\\
  {\small\texttt{giovanni.manzini@unipi.it}}
  \and
  Guilherme P. Telles$^{2}$\\
  {\small\texttt{gpt@ic.unicamp.br}}
}

\date{\small
$^{1}$\,Faculdade de Engenharia El\'etrica, Universidade Federal de Uberl\^andia, Uberl\^andia, MG, Brazil\\
$^{2}$\,Instituto de Computa\c{c}\~ao, Universidade Estadual de Campinas, Campinas, SP, Brazil\\
$^{3}$\,Department of Computer Science, University of Pisa, Pisa, Italy}

\begin{document}

\maketitle

\begin{abstract}
\noindent
Iterative machine-learning algorithms repeatedly access large real-valued matrices through matrix--vector products, and their performance is limited by memory bandwidth rather than arithmetic throughput.
Lossless compression schemes that support computation directly over the compressed data reduce both the data moved and the memory footprint of these workloads.
The \mmRepair framework~\cite{FerraginaMGKNST22} follows this approach by applying the \Repair grammar compressor to the Compressed Sparse Row/Value (CSRV) representation of a matrix and performing left and right matrix--vector multiplications directly on the compressed data.
Its effectiveness, however, degrades on matrices that are dense or contain many distinct values, since their CSRV sequence has little repetition for a grammar to exploit.
In this paper we present a hybrid strategy for the \mmRepair framework that combines grammar and entropy compression.
The CSRV sequence is partitioned into a grammar-compressible component, formed by the entries that \Repair can compress effectively, and an entropy-compressible component holding the remaining entries, which is encoded with the \ansfold entropy coder~\cite{Moffat2020}.
Compression improves further by sorting and delta encoding the entries of each row in the CSRV representation, and by remapping the alphabet to a contiguous range whenever a simple cost model predicts a saving.
Experiments on real-valued machine-learning matrices, including word embeddings and transformer hidden states, show that our hybrid strategy is more compact than the original \mmRepair on every input and reduces the peak memory of multiplication, with the largest gains on dense matrices with many distinct values, at competitive multiplication times.
It also outperforms the column-reordering variant of \mmRepair on most of the matrices, and combining the two is even more compact on the sparse ones, at the cost of a much longer compression time.
Finally, our strategy scales with the number of threads as well as the original \mmRepair does.
\end{abstract}


\section{Introduction}\label{sec:introduction}

Machine learning (ML) has become one of the main consumers of the data stored and managed by information systems, and much of this data takes the form of real-valued matrices~\cite{Wang2020}.
A training set is a matrix with one row per observation and one column per feature, an embedding table has one row per token, and the activations of a neural network form a matrix with one row per input.
Such matrices can contain billions of entries, and the algorithms that process them are often iterative: gradient descent, conjugate gradients, power iteration, and their variants repeatedly access the matrix through matrix--vector products~\cite{Elgohary2018,Elgohary2019}.
Because a matrix--vector product performs only a small number of floating-point operations per nonzero entry, these workloads are often memory-bandwidth-bound, making data movement a major contributor to execution time and energy consumption~\cite{TosoniBBAFM25}.
Compression is therefore attractive because it reduces storage requirements and the amount of data transferred through the memory hierarchy, potentially improving both memory capacity and effective bandwidth.

Lossy techniques, such as quantization and sparsification, can degrade accuracy in ways that depend on the data and the learning algorithm.
This has motivated research on lossless matrix compression that preserves the original numerical values.
General-purpose compressors, such as \texttt{gzip} and \texttt{xz}, are poorly suited to matrix computations because they treat a matrix as a one-dimensional byte stream, without explicitly exploiting dependencies among rows and columns.
More importantly, the input must be fully decompressed before any linear-algebra operation can take place, so the total cost comprises both the decompression and the subsequent operation on the uncompressed matrix.

To overcome these limitations, the Compressed Linear Algebra (CLA) framework was proposed in~\cite{Elgohary2018,Elgohary2019}.
CLA applies compression to matrix representations and performs linear-algebra operations over the compressed data, enabling larger datasets to fit in memory while avoiding full decompression.
More recently, Ferragina \etal~\cite{FerraginaMGKNST22} introduced \mmRepair, which applies the \Repair grammar compressor~\cite{Larsson2000} to a Compressed Sparse Row/Value (CSRV) representation of the input matrix.
The resulting representation supports both left and right matrix--vector multiplications directly on the compressed data, with time and memory proportional to the compressed size.
Moreover, the compression achieved by \mmRepair is theoretically bounded in terms of the $k$-th order empirical entropy of the CSRV representation.
Experiments have shown that \mmRepair outperforms CLA in both space and computation time on real-world ML matrices.
The recent survey of Navarro~\cite{Navarro26survey} places \mmRepair among the few computation-friendly formats that operate directly on real-valued matrices.

The effectiveness of \mmRepair, however, depends on the repetitiveness of the CSRV representation.
It performs particularly well on sparse matrices and on matrices with relatively few distinct nonzero values, characteristics commonly found in many ML workloads.
On denser matrices and on matrices with many distinct values, such as word embeddings~\cite{pennington-etal-2014-glove} and neural-network hidden states~\cite{yu2023compressing}, its compression effectiveness decreases.
In these cases, the CSRV sequence has a large terminal alphabet and few repeated patterns for grammar compression to exploit.
Furthermore, \Repair replaces only symbol pairs that occur at least twice, leaving low-frequency symbols as terminals.
When such symbols account for a substantial fraction of the sequence, grammar compression provides limited savings over the original CSRV representation.
These limitations motivate a hybrid approach that combines grammar compression with an entropy coder better suited to the entries that offer little benefit to the grammar.

\paragraph{Our contribution.}
We introduce a hybrid compression strategy for the \mmRepair framework, called \our, that addresses the above limitations.
A preliminary version of this work appeared in~\cite{LouzaMT2027}.
The input matrix is partitioned into two complementary components, $M=M_A+M_B$, where $M_A$ contains the entries suited to grammar compression and $M_B$ the remaining ones.
The CSRV representation of $M_A$ is compressed with \Repair as in \mmRepair, while that of $M_B$ is encoded with the entropy compressor \ansfold~\cite{Moffat2020}, which is designed for large alphabets.
Since the entries within a CSRV row can be arbitrarily reordered without affecting matrix--vector multiplication, we further improve entropy compression through row-wise sorting and delta encoding.
Finally, when the expected savings justify it, we remap the CSRV alphabet to a compact contiguous range, reducing the space required to represent terminal symbols.
Matrix--vector multiplication is performed independently on the two components and combined as $Mx = M_Ax + M_Bx$, so the representation retains the key property of \mmRepair, that no decompression is needed, and we give its time and space costs in terms of the sizes of the two components.
Experimental results on real-valued machine-learning matrices show that \our is more compact than \mmRepair on all inputs and reduces peak memory during matrix--vector multiplication, with the largest gains on denser matrices and on matrices with many distinct values, while multiplication times remain competitive.
A component analysis isolates the contribution of each technique, and a study on quantized word embeddings relates the gains to the number of distinct values.
\our also outperforms the column-reordering variant of \mmRepair, which permutes the columns before compression, on most of the matrices.
The combination of \our with column reordering is more compact still on the sparse matrices, at the cost of a much longer compression time due to the reordering.
Finally, in a multithreaded setting with the matrix divided into row blocks, one per thread, \our attains the same speedups as \mmRepair and keeps its lower peak memory at every thread count.

\section{Preliminaries}\label{sec:background}


\subsection{Grammar compression and \Repair}\label{sec:repair}

Grammar compression represents a string $T$ over an alphabet $\Sigma$ using a context-free grammar that generates only $T$~\cite{CharikarLLPPSS05}.
We consider grammars in the form of a \emph{straight-line program} (SLP)~\cite{Lohrey12}, consisting of $q$ production rules $X_i\rightarrow AB$, one for each nonterminal $X_i$, where $A$ and $B$ are terminals or nonterminals.
The rules are numbered so that if $X_i$ occurs in the right-hand side of a rule for $X_j$, then $i<j$.
The \emph{expansion} $\mathit{exp}(X_i)$ of a nonterminal $X_i$ is the string over $\Sigma$ obtained by recursively applying the production rules.
A grammar-compressed representation consists of the rule set $\mathcal{R}$ and a final sequence $\mathcal{C}$ of terminals and nonterminals whose expansion yields $T$.
A substring that occurs many times in $T$ is thus stored once, as the expansion of a nonterminal, and referenced by one symbol at each occurrence.
We measure the size of a grammar-compressed representation by the total number of stored symbols, $|\mathcal{R}|+|\mathcal{C}|$, where $|\mathcal{R}|=2q$ counts two symbols per rule.
This measure determines the time complexity of matrix--vector multiplication in \mmRepair and is used throughout the paper.

\Repair (Recursive Pairing)~\cite{Larsson2000} is a well-known grammar compressor that iteratively selects the most frequent adjacent pair of symbols (\emph{digram}) $AB$ in the current sequence, replaces all its non-overlapping occurrences with a new nonterminal $X$, and adds the rule $X\rightarrow AB$ to $\mathcal{R}$.
The process stops when every digram occurs at most once. The remaining sequence of terminals and nonterminals forms the final sequence $\mathcal{C}$.
RePair can be implemented in linear time and space.
Its replacement rule provides the basis for the partitioning in Section~\ref{sec:split}: digrams occurring at most twice yield no immediate reduction in the number of stored symbols, while more frequent digrams offer opportunities for grammar compression.

Finding the smallest grammar for a string is NP-hard~\cite{CharikarLLPPSS05}.
Nevertheless, \Repair and other compressors that produce irreducible grammars achieve a representation of at most $|T|H_k(T)+o(|T|H_k(T))$ bits for every $k\in o(\log_{|\Sigma|}|T|)$, where $H_k(T)$ denotes the $k$-th order empirical entropy of $T$~\cite{KiefferY00,OchoaN19}.
Thus, up to lower-order terms, these compressors achieve the same asymptotic space as a statistical coder that models each symbol using its $k$ preceding symbols.

\subsection{The CSRV representation}\label{sec:csrv}

Given a matrix $M\in\mathbb{R}^{n\times m}$ with $t$ nonzero entries, the classic compressed sparse row (CSR) representation~\cite{Saad2003} stores the nonzeros row by row together with their column indices and the starting position of each row.
When the number of distinct nonzero values is small, it is more economical to keep the distinct values in an array $V$ and to store, for each nonzero, its index in $V$ rather than the value itself, which is the idea of the CSR-IV format~\cite{KourtisGK08}.
The Compressed Sparse Row/Value (CSRV) representation~\cite{FerraginaMGKNST22} combines this idea with a single-sequence layout.
It consists of the array $V$ of the distinct nonzero values of $M$ and of a sequence $S$ of length $t+n$ obtained by scanning $M$ row by row, appending for each nonzero $M[i,j]$ the pair $\langle\ell,j\rangle$ such that $V[\ell]=M[i,j]$, and appending a delimiter symbol $\$$ at the end of each row.
Each symbol of $S$ is either a pair $\langle\ell,j\rangle$, with $0\leq\ell<|V|$ and $0\leq j<m$, or $\$$, and the delimiters partition $S$ into $n$ row sequences, which replace the array of row pointers of CSR.
Figure~\ref{fig:csrv} shows an $8\times 6$ matrix and its CSRV representation.
Storing the association between values and their index in $V$ in a hash table, the representation is built in $O(nm)$ expected time~\cite[Lemma~2.1]{FerraginaMGKNST22}.
The single-sequence layout is what makes grammar compression applicable to a matrix, since the same pair of symbols recurring in many rows becomes a repeated digram of $S$.

\begin{figure}[t]
\centering
\begin{tikzpicture}[font=\small,
  idx/.style={fill=black!12, inner xsep=3pt, inner ysep=1.5pt, font=\scriptsize, text=black!70, minimum width=2mm},
  sub/.style={font=\small\bfseries, anchor=north}]

\matrix (M) [matrix of math nodes, ampersand replacement=\&, left delimiter={[}, right delimiter={]},
  nodes={inner xsep=4pt, inner ysep=2pt, anchor=center, minimum width=7mm},
  column sep=1pt, row sep=2pt] at ($(-1.0,-2.0)$)
{
5.3 \& 0   \& 8.1 \& 8.1 \& 6.0 \& 5.3 \\
5.3 \& 0   \& 0   \& 2.7 \& 0   \& 6.0 \\
5.3 \& 0   \& 8.1 \& 0   \& 6.0 \& 5.3 \\
6.0 \& 0   \& 0   \& 2.7 \& 0   \& 6.0 \\
0   \& 0   \& 0   \& 0   \& 6.0 \& 5.3 \\
2.7 \& 0   \& 6.0 \& 2.7 \& 5.3 \& 0   \\
0   \& 0   \& 6.0 \& 8.1 \& 0   \& 8.1 \\
2.7 \& 0   \& 6.0 \& 2.7 \& 5.3 \& 0   \\
};
\node[anchor=east] at ($(M.west)+(-0.45,0)$) {$M=$};
\foreach \c [evaluate=\c as \j using int(\c-1)] in {1,...,6}
  \node[idx] at ($(M-1-\c.north)+(0,0.42)$) {$\j$};

\matrix (V) [matrix of math nodes, ampersand replacement=\&, left delimiter={[}, right delimiter={]},
  nodes={inner xsep=4pt, inner ysep=2pt, anchor=center, minimum width=7mm},
  column sep=0pt, row sep=0pt, anchor=north west] at ($(M.north east)+(2.4,0.5)$)
{
5.3 \& 8.1 \& 6.0 \& 2.7 \\
};
\node[anchor=east] at ($(V.west)+(-0.45,0)$) {$V=$};
\foreach \c [evaluate=\c as \l using int(\c-1)] in {1,...,4}
  \node[idx] at ($(V-1-\c.north)+(0,0.42)$) {$\l$};
 
\node[anchor=north west, inner sep=0pt] (S) at ($(V.west)+(-1.32,-0.55)$) {%
$\begin{aligned}
S ={}& \langle 0,0\rangle\langle 1,2\rangle\langle 1,3\rangle\langle 2,4\rangle\langle 0,5\rangle\$\,\\
     &   \langle 0,0\rangle\langle 3,3\rangle\langle 2,5\rangle\$ \\
     & \langle 0,0\rangle\langle 1,2\rangle\langle 2,4\rangle\langle 0,5\rangle\$\,\\ &
        \langle 2,0\rangle\langle 3,3\rangle\langle 2,5\rangle\$\,
        \langle 2,4\rangle\langle 0,5\rangle\$ \\
     & \langle 3,0\rangle\langle 2,2\rangle\langle 3,3\rangle\langle 0,4\rangle\$\, \\&
        \langle 2,2\rangle\langle 1,3\rangle\langle 1,5\rangle\$ \\
     & \langle 3,0\rangle\langle 2,2\rangle\langle 3,3\rangle\langle 0,4\rangle\$
\end{aligned}$};
 
\end{tikzpicture}
\caption{A matrix $M$ (left) and its CSRV representation $(V,S)$ (right).
The array $V$ holds the distinct nonzero values of $M$, and $S$ is a single sequence of $t+n$ symbols, obtained by scanning $M$ row by row: each nonzero $M[i,j]$ becomes the pair $\langle\ell,j\rangle$ with $V[\ell]=M[i,j]$, and the delimiter $\$$ closes each row.
In $S$ the pair $\langle 2,5\rangle$ of the second row denotes the value $V[2]=6.0$ in column $5$, while the same value in column $4$ of the first row is the different pair $\langle 2,4\rangle$, since a pair identifies a value in a given column.}
\label{fig:csrv}
\end{figure}

Two properties of $S$ are used throughout the paper.
First, matrix--vector products can be computed with a single scan of $S$: for $y=Mx$ one adds $V[\ell]\cdot x[j]$ to $y[i]$ for every pair $\langle\ell,j\rangle$ of row $i$, and for $x^T=y^TM$ one adds $y[i]\cdot V[\ell]$ to $x[j]$, so both products take $O(n+t)$ time.
Second, since every pair carries its column index, the symbols within a row can be arbitrarily reordered without changing the represented matrix or its products~\cite[Sect.~3]{FerraginaMGKNST22}.
For example, the first row of $S$ in Figure~\ref{fig:csrv}, $\langle 0,0\rangle\langle 1,2\rangle\langle 1,3\rangle\langle 2,4\rangle\langle 0,5\rangle$, can be written as $\langle 1,3\rangle\langle 0,0\rangle\langle 1,2\rangle\langle 2,4\rangle\langle 0,5\rangle$.
The order matters for compression, however: in the new order the third row, $\langle 0,0\rangle\langle 1,2\rangle\langle 2,4\rangle\langle 0,5\rangle$, occurs entirely within the first one and can be represented by a single nonterminal.
This freedom is exploited by the column reordering of \mmRepair (Section~\ref{sec:mmrepair}) and by the row ordering of Section~\ref{sec:ole}.

\subsection{\mmRepair}\label{sec:mmrepair}

\begin{figure}[t]
\centering
\fitwidth{%
\includegraphics[width=\textwidth]{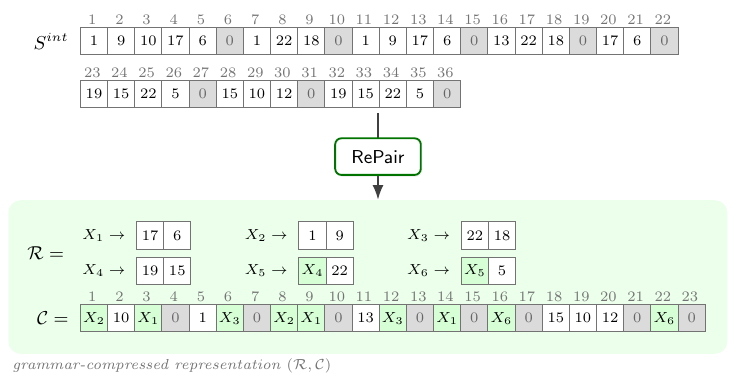}
}
\caption{The \mmRepair representation of the matrix of Figure~\ref{fig:csrv}.
The top part shows the integer sequence $S^{int}$ with $m=6$, obtained by encoding $\$$ as $0$ and each pair $\langle\ell,j\rangle$ as $\ell m+j+1$, and the numbers above the cells are positions.
The bottom part shows the rule set $\mathcal{R}$ and the final sequence $\mathcal{C}$ produced by \Repair.
Since no rule may contain $0$, every delimiter remains in $\mathcal{C}$, which also contains terminals, such as $10$ in the first row and $1$ in the second.}
\label{fig:mmrepair}
\end{figure}

The \mmRepair framework, introduced by Ferragina \etal~\cite{FerraginaMGKNST22}, applies \Repair to the CSRV sequence $S$.
Before compression, $S$ is converted into an integer sequence $S^{int}$, in which the delimiter $\$$ is encoded as $0$ and each pair $\langle\ell,j\rangle$ as the integer $\ell m+j+1$, where $m$ is the number of columns of $M$.
Because $0\leq j<m$, the pair is recovered from its identifier by Euclidean division.
We denote by $\Sigma$ the alphabet of $S^{int}$, and when the encoding is not relevant we keep writing $\langle\ell,j\rangle$ and $\$$ for its symbols.
\Repair is then modified so that no rule contains the delimiter $0$.
This guarantees that the expansion of every nonterminal is a sequence of pairs belonging to one row and that all delimiters remain explicit in the final sequence $\mathcal{C}$.
A representation in which each row is represented by a single nonterminal would have $\mathcal{C}=X_{i_1}\$X_{i_2}\$\cdots X_{i_n}\$$, with $\mathit{exp}(X_{i_r})$ equal to row $r$.
However, as \Repair stops when every digram occurs at most once, the final sequence $\mathcal{C}$ may contain multiple terminals and nonterminals per row.
The multiplication algorithms therefore handle both symbol types.
The resulting representation of $M$ is the triple $(\mathcal{R},\mathcal{C},V)$, shown in Figure~\ref{fig:mmrepair} for the matrix in Figure~\ref{fig:csrv}.
The grammar-compressed representation of $S$ is bounded in size in terms of its $k$-th order empirical entropy, providing the theoretical guarantee underlying \mmRepair.

The compression phase of \mmRepair takes $O(nm)$ expected time and $O(|S|)$ words of space: the CSRV representation is built in $O(nm)$ expected time (Section~\ref{sec:csrv}), and \Repair runs in time and space linear in $|S|$~\cite{Larsson2000}.
The subsequent encoding of $\mathcal{R}$ and $\mathcal{C}$, described below, takes time linear in their size.

The key property of the representation is that it supports left and right multiplication without decompression, in time proportional to its size.
For a vector $x\in\mathbb{R}^m$ let $\mathit{eval}_x(\langle\ell,j\rangle)=V[\ell]\cdot x[j]$ and, for a nonterminal $X_i$ with expansion $\langle\ell_1,j_1\rangle\cdots\langle\ell_h,j_h\rangle$, let $\mathit{eval}_x(X_i)=\sum_{k=1}^{h}V[\ell_k]\cdot x[j_k]$.
Then $\mathit{eval}_x(X_i)=\mathit{eval}_x(A)+\mathit{eval}_x(B)$ for every rule $X_i\rightarrow AB$, and if $y=Mx$ then $y[r]$ is the sum of $\mathit{eval}_x$ over the symbols of the $r$-th row of $\mathcal{C}$~\cite[Lemmas~3.2 and~3.3]{FerraginaMGKNST22}.

\begin{theorem}[\cite{FerraginaMGKNST22}, Theorems~3.4 and~3.10]
\label{thm:mmrepair}
Given the representation $(\mathcal{R},\mathcal{C},V)$ of a matrix $M\in\mathbb{R}^{n\times m}$, the products $y=Mx$ and $x^T=y^TM$ can be computed in $O(|\mathcal{R}|+|\mathcal{C}|)$ time using $O(|\mathcal{R}|)$ words of auxiliary space.
\end{theorem}

For $y=Mx$ the algorithm fills an array $W[1,q]$ with $W[i]=\mathit{eval}_x(X_i)$ by a single forward pass over $\mathcal{R}$, which is possible because the two symbols of each rule are either terminals or nonterminals with a smaller index.
It then scans $\mathcal{C}$, summing $\mathit{eval}_x$ of terminals and $W$ of nonterminals into the current entry of $y$.
For $x^T=y^TM$ the roles are reversed: $W[i]$ accumulates $\mathit{sum}_y(X_i)$, the sum of $y[r]$ over the rows $r$ whose compressed representation uses $X_i$.
The entries of $W$ are initialized from the nonterminals occurring in $\mathcal{C}$, the rules are scanned backwards from $X_q$ to $X_1$, and for each $X_j\rightarrow AB$ the value $W[j]$ is added to $W[i]$ if $A$ or $B$ is a nonterminal $X_i$, and $V[\ell]\cdot W[j]$ is added to $x[k]$ if $A$ or $B$ is a terminal $\langle\ell,k\rangle$.
Two aspects of this algorithm constrain the representation: the rule set must support a backward sequential scan, and the final sequence a forward one.
We refer the reader to~\cite{FerraginaMGKNST22} for the proofs.
The auxiliary array $W$, with one double per rule, is the reason why the peak memory of a multiplication exceeds the compressed size, and why a smaller grammar reduces both, a point we return to in Section~\ref{sec:mult}.

\begin{figure}[t]
\centering
\includegraphics[width=0.95\textwidth]{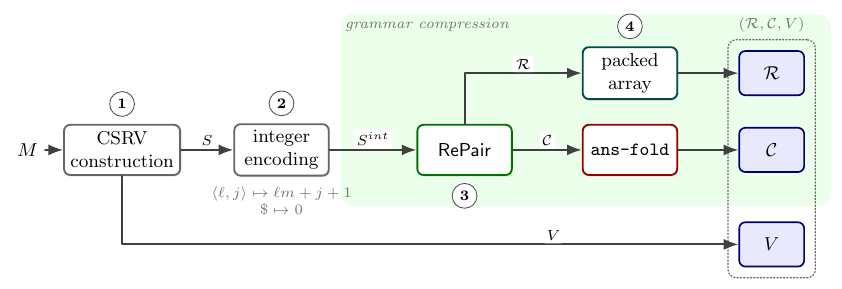}
\caption{
Compression pipeline of \mmRepair in its \reans variant.
The matrix is converted into its CSRV representation $(V,S)$ (step~1), the sequence $S$ is encoded as the integer sequence $S^{int}$ (step~2), $S^{int}$ is compressed with \Repair, modified so that no rule contains the delimiter $0$, into the rule set $\mathcal{R}$ and the final sequence $\mathcal{C}$ (step~3), and $\mathcal{R}$ is stored as a packed array while $\mathcal{C}$ is entropy coded with \ansfold (step~4).
The compressed representation is the triple $(\mathcal{R},\mathcal{C},V)$, with $V$ stored uncompressed.}
\label{fig:pipeline-mmrepair}
\end{figure}

In the implementation of \mmRepair, summarized in Figure~\ref{fig:pipeline-mmrepair}, $S^{int}$ is stored as an array of 32-bit unsigned integers.
Three representations of $(\mathcal{R},\mathcal{C})$ were tested in~\cite{FerraginaMGKNST22}, with different space--time trade-offs: \texttt{re-32} stores both as 32-bit integers and is the fastest, \texttt{re-iv} stores both as packed arrays, built with the \texttt{sdsl-lite} library~\cite{Gog2014a}, with \textcolor{blue}{$1+\lfloor\log_2 N_{\max}\rfloor$} bits per symbol, where $N_{\max}$ is the largest symbol identifier, and \reans keeps $\mathcal{R}$ as a packed array and compresses $\mathcal{C}$ with \ansfold~\cite{Moffat2020}, a semi-static entropy coder for integer sequences over large alphabets based on asymmetric numeral systems.
In all of them $V$ is stored as uncompressed 8-byte doubles.
\reans is the most compact of the three and the one shown to be more compact and faster than CLA on real ML matrices~\cite{FerraginaMGKNST22}, and more energy-efficient than other compressed formats in the study of~\cite{TosoniBBAFM25}.
For multithreaded execution, \mmRepair divides the matrix into row blocks that are compressed independently and share $V$, a scheme we keep and discuss in Section~\ref{sec:parallel}.

\paragraph{Column reordering.}
Since the symbols of a row can be listed in any order (Section~\ref{sec:csrv}), Ferragina \etal~\cite[Sect.~5]{FerraginaMGKNST22} also permute the columns of $M$ so that columns whose values frequently occur together become adjacent and form digrams that \Repair can replace.
The order should maximize the total similarity of consecutive columns, measured by their repeated pairs of nonzero values, a variant of the traveling salesman problem.
The best trade-off in~\cite{FerraginaMGKNST22} is given by \texttt{PathCover}, which greedily covers the similarity graph with vertex-disjoint paths of maximum weight, on a similarity matrix pruned to the $k$ most similar columns of each column, with one permutation per row block.
The permutation need not be stored, since every pair keeps its column index, but building the similarity matrix takes time quadratic in the number of columns.

\section{Hybrid \mmRepair}\label{sec:hybrid}

The effectiveness of \Repair depends on the presence of repeated digrams.
Symbols that occur only once, or only in rarely occurring digrams, offer little opportunity for compression and remain as terminals in the final sequence $\mathcal{C}$.
Motivated by this observation, we propose a hybrid compression strategy that partitions the CSRV sequence $S$ into two components: the grammar-compressible component $S_A$, which is compressed using the \mmRepair pipeline, and the remaining component $S_B$, which is compressed independently using the \ansfold entropy coder.
This partition induces a decomposition $M=M_A+M_B$, where $M_A$ and $M_B$ contain the entries represented by $S_A$ and $S_B$, respectively.

\subsection{Partitioning the CSRV sequence}\label{sec:split}

The partitioning rule is driven by the following accounting of the effect of a single \Repair rule on the number of stored symbols.

\begin{proposition}\label{prop:digram}
Let $AB$ be a digram with $k\geq 2$ non-overlapping occurrences in the current sequence.
Replacing all of them with a new nonterminal $X$ and adding the rule $X\rightarrow AB$ reduces the total number of stored symbols $|\mathcal{R}|+|\mathcal{C}|$ by $k-2$.
Thus, the replacement leaves the representation size unchanged when $k=2$ and reduces it when $k\geq 3$.
\end{proposition}
\begin{proof}
The $k$ occurrences account for $2k$ symbols in the sequence, which are replaced by $k$ occurrences of $X$, saving $k$ symbols.
The rule $X\rightarrow AB$ adds two symbols to $\mathcal{R}$.
Therefore, the total reduction in the number of stored symbols is $k-2$.
\end{proof}

Proposition~\ref{prop:digram} ignores the effect that a replacement has on the frequencies of other digrams (a new nonterminal may itself participate in frequent digrams), so it is not a guarantee about the final grammar.
It does, however, identify the entries that can offer no direct saving, since the frequency of a digram of original symbols can only decrease as \Repair proceeds.
This leads to the following rule.

\begin{definition}[Partitioning rule]\label{def:split}
Let $f(AB)$ denote the number of occurrences of the digram $AB$ in $S$.
An entry $S[i]\neq\$$ is assigned to $S_A$ if at least one of its incident digrams satisfies $f(S[i-1]S[i])>2$ or $f(S[i]S[i+1])>2$, where digrams containing the delimiter $\$$ are ignored.
All remaining entries are assigned to $S_B$.
The delimiter closing each row is appended to both $S_A$ and $S_B$.
\end{definition}

The rule captures three cases.
If $S[i]$ occurs only once in $S$, it cannot participate in any repeated digram and is therefore uncompressible by \Repair: it would remain as a terminal symbol in $\mathcal{C}$.
If $S[i]$ is not unique but each of its incident digrams occurs only once, neither participates in a grammar rule, and $S[i]$ would likewise remain as a terminal in $\mathcal{C}$.
Finally, if neither incident digram occurs more than twice, by Proposition~\ref{prop:digram} no rule built on those digrams can reduce the number of stored symbols $|\mathcal{R}|+|\mathcal{C}|$, and we regard $S[i]$ as ineffective for grammar compression.
In all three cases, $S[i]$ is assigned to $S_B$.

These are heuristic choices: they identify entries of little immediate benefit to \Repair, whose removal may expose new repetitions that grammar compression can exploit.
For example, removing the unique symbol $x$ from $A,C,A,C,A,x,C,A,C$ yields $A,C,A,C,A,C,A,C$, where the digram $AC$ occurs four times instead of three.
The rule is also deliberately local: it looks only at the two digrams incident to each entry, whose frequencies are collected in a single pass over $S$, whereas deciding the exact benefit of an entry for the final grammar would require running \Repair itself.

\begin{lemma}\label{lem:split-time}
The partition of Definition~\ref{def:split} can be computed with two passes over $S$ using $O(d)$ words of space, where $d\leq|S|$ is the number of distinct digrams of $S$, in $O(|S|)$ expected time with a hash table.
\end{lemma}
\begin{proof}
A first pass over $S$ stores the frequency of every digram not containing $\$$ in a dictionary indexed by the pair of consecutive symbols, which holds $d$ entries.
A second pass looks up the at most two incident digrams of every entry and appends it to $S_A$ or $S_B$.
Each pass performs $O(|S|)$ dictionary operations, each of which takes $O(1)$ expected time in a hash table.
\end{proof}

Since every nonzero entry of $M$ belongs to exactly one component, while the delimiter symbols are included in both, the two components satisfy the following relation.

\begin{lemma}\label{lem:size}
Let $n$ be the number of rows of $M$ and $t$ its number of nonzeros.
Then $|S_A|+|S_B| = |S|+n = t+2n$, and each of $S_A$ and $S_B$ contains exactly $n$ delimiters.
\end{lemma}

\begin{figure}[t]
\centering
\includegraphics[width=\textwidth]{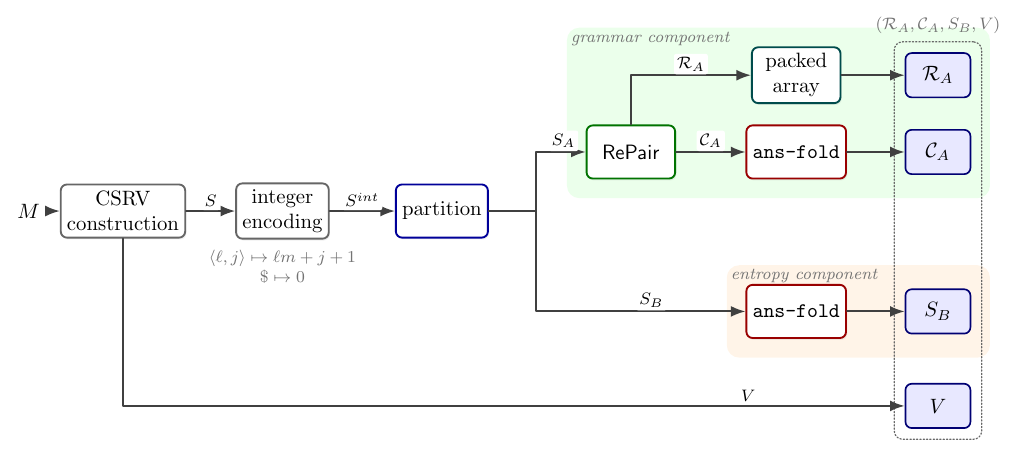}
\caption{Overview of the partitioning scheme.
The integer sequence $S^{int}$ is partitioned into two components: $S_A$ is compressed with \Repair as in \mmRepair, and $S_B$ is encoded with \ansfold, while $V$ is stored unchanged.}
\label{fig:hybrid-overview}
\end{figure}

The \mmRepair pipeline is applied to $S_A$, obtaining a grammar consisting of the rules $\mathcal{R}_A$ and the final sequence $\mathcal{C}_A$, while $S_B$ is compressed using \ansfold.
The value array $V$ remains unchanged from the original \mmRepair representation.
The compressed representation of $M$ is thus the tuple $(\mathcal{R}_A,\mathcal{C}_A,S_B,V)$, summarized in Figure~\ref{fig:hybrid-overview}.

\subsection{Row reordering and delta encoding}\label{sec:ole}

We further exploit the fact that the nonzero entries within each row of a CSRV sequence
can be reordered without changing the represented matrix (Section~\ref{sec:csrv}).
Since in $S^{int}$ each pair is represented by the integer identifier $\ell m+j+1$, the identifiers of a row can be sorted in increasing order and delta encoded: the first identifier is stored unchanged, and each subsequent one is replaced by its difference from the previous one, which produces small integers that an entropy coder represents compactly.
We apply this idea to $S_B$ and to $\mathcal{C}_A$ before encoding them.
In $S_B$ every row is a sequence of terminals.
In $\mathcal{C}_A$ a row contains terminals and nonterminals, and since \Repair assigns to the nonterminals identifiers larger than those of the terminals, we sort both kinds together.
We refer to this transformation as ordered list encoding (\textsc{ole}).




For example, the first row of $S^{int}$ in Figure~\ref{fig:mmrepair}, $1,9,10,17,6$, becomes $1,6,9,10,17$ after sorting and $1,5,3,1,7$ after delta encoding, so that no value exceeds 7, against 17 before.

The transformation turns every row into the gap encoding of an increasing sequence, the classical representation of inverted lists~\cite{Navarro2016}.
If a row has $r$ distinct identifiers from $\{1,\ldots,\sigma\}$, its gaps $g_1,\ldots,g_r$ satisfy $\sum_k\log_2 g_k\leq r\log_2(\sigma/r)$, so a code that spends about $\log_2 g$ bits on a gap $g$ encodes the row in about $r\log_2(\sigma/r)$ bits, close to the information-theoretic minimum $\log_2\binom{\sigma}{r}$\cite[Sect.~2.3.1]{Navarro2016}.
An unsorted row of $r$ identifiers costs instead about $r\log_2\sigma$ bits under any zero-order model, so the gain of \textsc{ole} grows with the row length.
Although \ansfold is not a gap code in this sense, it assigns short codewords to small integers and models their distribution, so in practice it benefits from the small gaps in the same way.
The experimental component analysis of Section~\ref{sec:components} quantifies the effect.

\subsection{Alphabet remapping}\label{sec:map}

We also consider remapping the terminal alphabet of the CSRV representation to a compact contiguous range, and apply it only when the expected savings outweigh the cost of the reverse mapping, which must be stored and kept in memory during multiplication.
Since each pair $\langle\ell,j\rangle$ is encoded as $\ell m+j+1$, the alphabet $\Sigma$ of $S^{int}$ is potentially sparse and its symbols may require more space than necessary, since their width is determined by the largest integer occurring in $S^{int}$ rather than by $|\Sigma|$.

The remapping process replaces each terminal identifier by its rank in the sorted alphabet, producing a dense alphabet $\Sigma'=\{1,\ldots,|\Sigma|\}$ and preserving the delimiter identifier $0$.
The reverse mapping, called \wcode, stores, for each rank $r\in\Sigma'$, the original identifier of $\Sigma$.
Since it is the sorted alphabet, it is itself an increasing sequence and can be stored delta encoded and entropy coded, like the rows of Section~\ref{sec:ole}.
During multiplication, however, it must be available in decoded form to recover $\langle\ell,j\rangle$ in $O(1)$ time, and therefore costs $O(|\Sigma|)$ words of memory.

Alphabet remapping reduces the width of terminal identifiers but introduces an additional memory cost during matrix--vector multiplication: the reverse mapping must be decoded and kept in memory to recover the original identifiers. We therefore decide whether to apply remapping by comparing the potential savings in identifier storage with the cost of the decoded mapping. Let $|S^{\mathrm{int}}|$ denote the length of the CSRV sequence and let $b=\lceil\log_2(|\Sigma|+1)\rceil$ be the number of bits required to represent an identifier in $\Sigma'\cup\{0\}$. 
Since the original identifiers are stored as 32-bit integers (Section~\ref{sec:impl}), reducing their width to $b$ bits offers an estimated saving of $|S^{int}|(32-b)$ bits, whereas the decoded mapping takes $32\,|\Sigma|$ bits.
We therefore apply remapping only when the saving exceeds this cost, that is, when
\begin{equation}\label{eq:map}
|S^{int}|\,(32-b) > 32\,|\Sigma| .
\end{equation}
Otherwise, the original identifiers are retained, avoiding the additional memory overhead during multiplication.

Since this comparison is based on the uncompressed CSRV sequence, it does not predict the exact change in the final compressed size. The actual benefit also depends on how RePair and ans-fold encode the remapped identifiers. The above criterion is simple to compute and proved effective in our experiments (Section~\ref{sec:components}).


\subsection{The complete compression pipeline}\label{sec:pipeline}

\begin{figure}[t]
\centering
\includegraphics[width=1.03\textwidth]{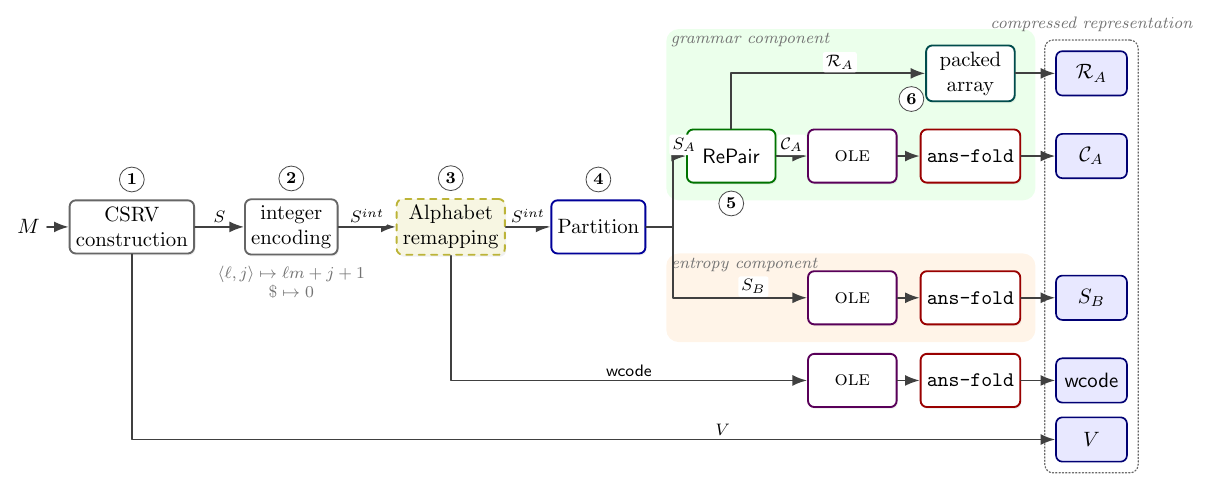}
\caption{Compression pipeline of hybrid \mmRepair.
The matrix is converted into its CSRV representation $(V,S)$ (step~1), and $S$ is encoded as the integer sequence $S^{int}$ (step~2).
When criterion~\eqref{eq:map} holds, the identifiers of $S^{int}$ are replaced by their ranks and the reverse mapping \wcode is produced (step~3).
The remapping is computed once on the whole sequence, so a single \wcode serves both components, and it does not affect the partition, which depends only on digram frequencies.
The sequence is partitioned into $S_A$ and $S_B$ (step~4), and $S_A$ is compressed with \Repair into $\mathcal{R}_A$ and $\mathcal{C}_A$ (step~5).
The rows of $\mathcal{C}_A$ and of $S_B$ are sorted and delta encoded (\textsc{ole}) and entropy coded with \ansfold, as is \wcode, while $\mathcal{R}_A$ is stored as a packed array (step~6).
The five blocks on the right form the compressed representation $(\mathcal{R}_A,\mathcal{C}_A,S_B,V)$, plus \wcode.}
\label{fig:pipeline}
\end{figure}

\begin{figure}[!htb]
\centering
\fitwidth{%
\includegraphics[width=0.9\textwidth]{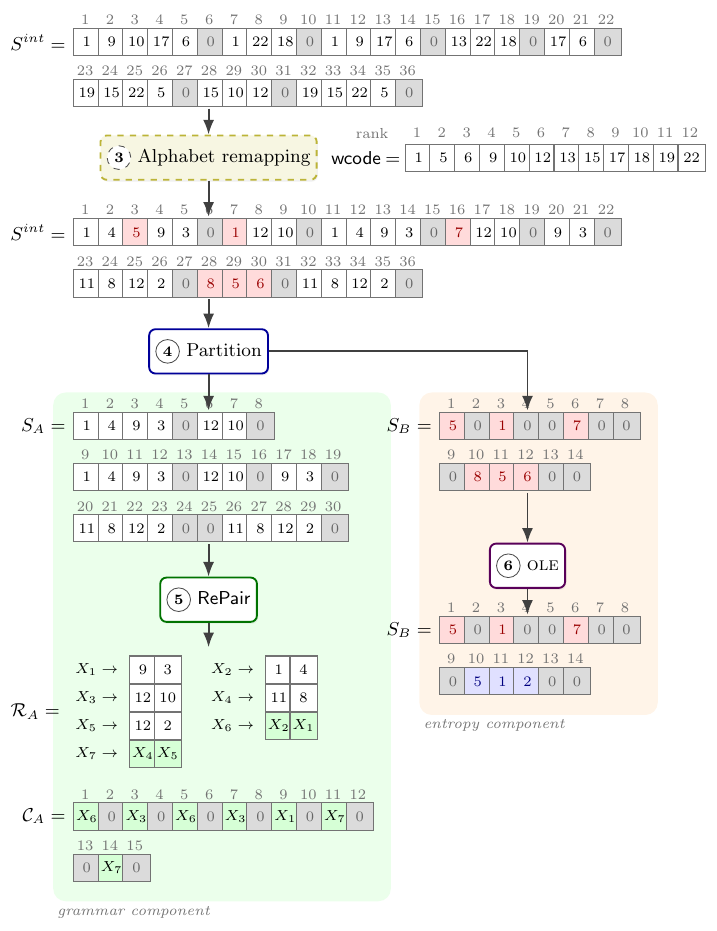} 
}
\caption{
The compression pipeline of hybrid \mmRepair on the matrix of Figure~\ref{fig:csrv}.
To keep the example small, an incident digram occurring twice is enough here to keep an entry in $S_A$.
The CSRV sequence is encoded as $S^{int}$ (steps~1--2).
Criterion~\eqref{eq:map} holds, since $|S^{int}|=36$, $|\Sigma|=12$ and $b=4$ give $36\cdot 28>32\cdot 12$, 
producing \wcode
(step~3).
The remapped sequence is partitioned into $S_A$ and $S_B$ (step~4), $S_A$ is compressed with \Repair (step~5), and the rows of $S_B$ are sorted and delta encoded (step~6).
}
\label{fig:split}
\end{figure}

The complete compression phase of hybrid \mmRepair combines the techniques of Sections~\ref{sec:split}--\ref{sec:map} as shown in Figure~\ref{fig:pipeline}, and Figure~\ref{fig:split} traces it on the running example.

\begin{proposition}\label{prop:compress-time}
The compression phase runs in $O(nm)$ expected time plus the time of \Repair on $S_A$ and of \ansfold on $\mathcal{C}_A$, $S_B$ and \wcode, and uses $O(|S|)$ words of working space in addition to that of \Repair.
\end{proposition}
\begin{proof}
Building $(V,S)$ takes $O(nm)$ expected time (Section~\ref{sec:csrv}), and the partition takes $O(|S|)$ expected time (Lemma~\ref{lem:split-time}).
The identifiers are integers smaller than $2^{30}$ (Section~\ref{sec:impl}), so both sorts can be done by radix sort with a constant number of passes.
The remapping sorts the distinct identifiers and rewrites the sequence in $O(|S|)$ time.
For \textsc{ole}, sorting all the symbols of a sequence by the pair (row, identifier) sorts every row in $O(|S|)$ time in total.
Since $|S|=t+n\leq nm+n$, all these steps take $O(nm)$ time.
\end{proof}

This is the same bound as for \mmRepair (Section~\ref{sec:mmrepair}), so the additional steps of the hybrid representation do not change the asymptotic cost of compression.
In practice, the additional running time is dominated by the computation of the digram frequencies, the alphabet remapping and the encoding of $S_B$ and $\mathcal{C}_A$, as the experiments in Section~\ref{sec:compression-time} show.
\Repair runs only on $S_A$, which is shorter than $S$, but this saving is smaller than the cost of these additional steps.


\subsection{Matrix--vector multiplication}\label{sec:mult}

Multiplication over the hybrid representation is performed independently on the two components and combined as $Mx=M_Ax+M_Bx$ and $y^TM=y^TM_A+y^TM_B$.
The products with $M_A$ are computed with the \mmRepair algorithms of Section~\ref{sec:mmrepair} over the representation $(\mathcal{R}_A,\mathcal{C}_A,V)$.
The products with $M_B$ reduce to the multiplication algorithm over CSRV sequences of Section~\ref{sec:csrv}, since $(S_B,V)$ is a CSRV representation of $M_B$, with the only difference that each row of $S_B$ must first be entropy decoded and then delta decoded.
The same holds for the rows of $\mathcal{C}_A$, so neither sequence is ever decompressed as a whole.
When the remapping is applied, \wcode is decoded once at the beginning, in $O(|\Sigma|)$ time and words, after which every terminal is resolved to its pair $\langle\ell,j\rangle$ in $O(1)$ time.

\begin{theorem}\label{thm:hybrid}
Given the representation $(\mathcal{R}_A,\mathcal{C}_A,S_B,V)$ of a matrix $M\in\mathbb{R}^{n\times m}$, the products $y=Mx$ and $x^T=y^TM$ can be computed in $O(|\mathcal{R}_A|+|\mathcal{C}_A|+|S_B|)$ time using $O(|\mathcal{R}_A|)$ words of auxiliary space, plus $O(|\Sigma|)$ words for the decoded \wcode when the remapping is applied.
If $k_A$ and $k_B$ denote the number of non-delimiter symbols in $\mathcal{C}_A$ and $S_B$, the time bound equals $O(|\mathcal{R}_A|+k_A+k_B+n)$.
\end{theorem}
\begin{proof}
The products with $M_A$ take $O(|\mathcal{R}_A|+|\mathcal{C}_A|)$ time and $O(|\mathcal{R}_A|)$ words by Theorem~\ref{thm:mmrepair}, since every nonterminal of $\mathcal{R}_A$ expands to pairs of one row and $\mathcal{C}_A$ contains the $n$ delimiters.
The products with $M_B$ take $O(|S_B|)$ time and $O(1)$ words (Section~\ref{sec:csrv}).
Delta decoding a row costs time linear in its length and $O(1)$ words, and adding the two results costs $O(n)$ or $O(m)$ time, which is within the bound since both sequences contain $n$ delimiters and every column with a nonzero appears in some symbol.
Finally, $|\mathcal{C}_A|+|S_B|=k_A+k_B+2n$ by Lemma~\ref{lem:size}.
When the remapping is applied, the decoded \wcode occupies $O(|\Sigma|)$ words and resolves every terminal in $O(1)$ time, so the time bound is unchanged.
\end{proof}

The representation itself occupies $|V|$ doubles for $V$, the packed array $\mathcal{R}_A$ and the compressed $\mathcal{C}_A$ and $S_B$, plus the $|\Sigma|$ words of the decoded \wcode when the remapping is applied.

Compared with the $O(|\mathcal{R}|+|\mathcal{C}|)=O(|\mathcal{R}|+k+n)$ time of \mmRepair, where $k$ is the number of non-delimiter symbols in $\mathcal{C}$, the hybrid representation pays an additional $O(n)$ term for the duplicated delimiters.
The grammar $\mathcal{R}_A$ is usually smaller than $\mathcal{R}$, since the entries moved to $S_B$ do not participate in rules, which reduces both the evaluation cost of $W$ and the auxiliary space.
On the other hand, the entries of $S_B$ are processed one at a time, whereas in $\mathcal{R}$ a nonterminal may cover many of them.
Which effect dominates depends on the input, and the experiments in Section~\ref{sec:running-time} show both cases.

\subsection{Parallel processing}\label{sec:parallel}

As in \mmRepair, the input matrix $M\in\mathbb{R}^{n\times m}$ is divided into $p$ row blocks of $\lceil n/p\rceil$ rows each, except possibly the last one, so that the $p$ blocks can be processed by $p$ threads.
The CSRV sequence $S$ is correspondingly divided into $p$ subsequences $S_1,\ldots,S_p$, which are compressed independently with the hybrid approach, while the value array $V$, the alphabet and the remapping decision remain global.
A right product then consists of $p$ independent block products, each producing a slice of $y$, while in a left product each block produces a partial vector of length $m$, and the $p$ vectors are summed at the end.
The bounds of Theorem~\ref{thm:hybrid} hold for each block, with an additional $O(pm)$ term for the final sum.

Because the blocks are compressed independently, repetitions spanning block boundaries cannot be exploited by \Repair.
The hybrid partitioning is affected as well, and in a predictable direction.

\begin{lemma}\label{lem:blocks}
Let $S_A^{(p)}$ denote the set of entries assigned to $S_A$ when Definition~\ref{def:split} is evaluated independently on each of $p$ row blocks, and $S_A^{(1)}$ the set obtained on the whole sequence.
Then $S_A^{(p)}\subseteq S_A^{(1)}$, and more generally $S_A^{(p')}\subseteq S_A^{(p)}$ whenever the $p'$ blocks refine the $p$ blocks.
\end{lemma}
\begin{proof}
The frequency of a digram within a block is at most its frequency within any block that contains it.
Hence, if an entry has an incident digram occurring more than twice in its block of the finer partition, the same digram occurs more than twice in its block of the coarser partition, and the entry is assigned to $S_A$ there as well.
\end{proof}

Consequently, increasing the number of blocks shifts entries from $S_A$ to $S_B$.
This may either increase the compressed size, because fewer repetitions are available to \Repair, or decrease it, when the entries moved to $S_B$ are encoded more compactly by \ansfold than by the grammar.
Section~\ref{sec:parallel-exp} shows both behaviors.


\subsection{Implementation details}\label{sec:impl}


Given a matrix $M\in\mathbb{R}^{n\times m}$, we first build its CSRV representation $(V,S)$ as described in Section~\ref{sec:csrv}.
As in \mmRepair, we store $S$ as an array of 32-bit unsigned integers, encoding $\$$ as $0$ and each pair $\langle\ell,j\rangle$ as $\ell m+j+1$ (identifiers of $2^{30}$ or greater are not supported), while the entries of $V$ are stored as 8-byte doubles.
Thus, the initial representation requires $4|S|+8|V|$ bytes.
The conversion tool reads $M$ row by row and builds $V$ using a hash table to identify distinct values.
It then computes the partition of Definition~\ref{def:split} using the two passes of Lemma~\ref{lem:split-time}, maintaining digram frequencies in a hash table, and stores the resulting sequences $S_A$ and $S_B$.
Finally, the tool evaluates criterion~\eqref{eq:map} on the alphabet of $S$ and, if the criterion is satisfied, replaces the identifiers in both sequences with their corresponding ranks in the sorted alphabet. The original identifiers are stored in $\wcode$, represented as the gaps between consecutive elements of the sorted alphabet.



To build the grammar component $(\mathcal{R}_A,\mathcal{C}_A)$ we compress $S_A$ with \texttt{irepair0}, the memory-bounded \Repair implementation used by \mmRepair, which takes the memory budget as a parameter and never builds a rule involving the delimiter $0$.
Terminals keep their identifiers and nonterminals receive the identifiers above them, so a symbol of $\mathcal{C}_A$ is a nonterminal if and only if it is at least the alphabet size, which \texttt{irepair0} stores as the first entry of the rule file.

In a na\"ive representation, the output of the compression phase consists of $|\mathcal{C}_A|+2|\mathcal{R}_A|+|S_B|$ 32-bit integers, plus $|\Sigma|$ integers for \wcode.
Each component is encoded according to the access pattern required by the multiplication algorithms, following the approach adopted in \reans.
The rule set $\mathcal{R}_A$ must be scanned forwards for right multiplication and backwards for left multiplication.
Since few compressors support efficient backward sequential access, $\mathcal{R}_A$ is stored in a packed array with $\lceil\log_2 N_{\max}\rceil$ bits per entry, where $N_{\max}$ is the largest symbol identifier, using the \texttt{int\_vector} class of \texttt{sdsl-lite}~\cite{Gog2014a}.
The left-hand sides of the rules are implicit, so a rule $X_i\rightarrow AB$ occupies entries $2i-1$ and $2i$.
In contrast, $\mathcal{C}_A$ and $S_B$ are accessed only sequentially in the forward direction, allowing us to use any encoding that supports sequential decoding.
We apply \textsc{ole} independently to each sequence: the identifiers within each row are sorted, and every identifier except the first is replaced by its difference from the preceding one, while the delimiter $0$ is preserved.
The resulting sequences are then compressed as single streams using \ansfold~\cite{Moffat2020}, the entropy coder used by \reans for $\mathcal{C}$, with fidelity parameter~1.
\ansfold groups integers into classes according to their magnitude, entropy-codes the class, and stores the remaining low-order bits explicitly.
Consequently, small integers, such as the gaps produced by \textsc{ole}, receive short codes.
The reverse mapping \wcode is also compressed with \ansfold, whereas $V$ is stored uncompressed.

For the multiplications, the compressed representations of $\mathcal{R}_A$, $\mathcal{C}_A$, $S_B$ and \wcode together with $V$ are loaded into memory.
Before the first product, \wcode is decoded once and turned into an array of $|\Sigma|$ 32-bit integers by a prefix sum of its gaps, after which a terminal $p$ is resolved as $\wcode[p]$, and $\langle\ell,j\rangle$ is recovered from $\wcode[p]-1$ by one Euclidean division by $m$.
The array $W$ holds $|\mathcal{R}_A|$ doubles.
The sequences $\mathcal{C}_A$ and $S_B$ are decoded into fixed buffers,
and the delta encoding is undone by a running sum reset at every delimiter.

\subsubsection{Multi-threaded implementation}\label{sec:impl-mt}

To take advantage of multi-core architectures, 
the input is partitioned into $p$ row blocks.
After computing $(V,S)$, the conversion tool writes the subsequence $S_i$ of each block to its own file, partitions it into its own $S_A$ and $S_B$, with the digram frequencies counted within the block, and evaluates criterion~\eqref{eq:map} once, on the alphabet of the whole $S$, so that $V$ and \wcode are unique and shared by all blocks.
The \Repair instances then run as parallel processes.
The compressed representation thus consists of $\mathcal{R}_A$, $\mathcal{C}_A$ and $S_B$ for each block, plus the global $V$ and \wcode.

For the multiplications, the program creates one POSIX thread per block, which lives for the whole run and is driven by the main thread through semaphores.
Each thread owns the structures of its block, namely its packed $\mathcal{R}_A$, its compressed $\mathcal{C}_A$ and $S_B$, its array $W$ and its buffers, while $V$, \wcode and the input vector are shared.
This per-thread state is why peak memory grows with $p$ (Section~\ref{sec:parallel-exp}).
For a right product each thread writes the slice of $y$ corresponding to its rows, and for a left product it accumulates into a private vector of length $m$, which the main thread sums at the end.


\renewcommand{\arraystretch}{1.2}

\section{Experiments}\label{sec:experiments}

We evaluate our implementation of the hybrid \mmRepair of Section~\ref{sec:hybrid}, called  \our, in terms of compression ratio, compression time, and matrix--vector multiplication performance.
We compare \our with \reans, the implementation of \mmRepair that achieved the best compression in~\cite{FerraginaMGKNST22}, and with \reansPC, the variant with \texttt{PathCover} column reordering described in Section~\ref{sec:mmrepair}.
We also report \ourPC, the combination of \our with \reansPC, introduced in Section~\ref{sec:pc}.
The original \mmRepair study showed that its compressed representation outperforms the CLA framework~\cite{Elgohary2018}, and therefore we do not include CLA in the comparison.
For reference, we also report the compression ratio and time of the general-purpose compressors \texttt{gzip} and \texttt{xz}, the latter with the option \texttt{-{}-lzma2=dict=128MiB}.
Recall that, unlike \mmRepair, both need to fully decompress the matrix before any operation can be performed.

All experiments were conducted on a server running Ubuntu 24.04.4 LTS (64-bit), equipped with an Intel Xeon Silver 4314 processor at 2.40\,GHz, 512\,GB of RAM, and 12\,TB of SATA storage.
Wall-clock execution times and peak memory were measured using \texttt{/usr/bin/time -v}.
The source code is available at \url{https://github.com/felipelouza/mm-repair-hybrid/}.

\subsection{Datasets}\label{sec:datasets}

\begin{table}[t]
\centering
\caption{Matrices used in our experiments.
The column \textit{nonzeros} reports the percentage of nonzero entries and \#distinct the number of distinct nonzero values.
The CSRV columns report \textit{maxval}, the largest identifier in $\Sigma$, and the alphabet size $|\Sigma|$.
The last column reports the fraction of $S$ assigned to $S_B$ by \our.}
\label{tab:datasets}
\small
\begin{tabular}{l|rr|rr|rr|r}
\hline
\multirow{2}{*}{Matrix}
& \multicolumn{2}{c|}{Dimensions}
& \multicolumn{2}{c|}{Density}
& \multicolumn{2}{c|}{CSRV} & Partition \\
\cline{2-8}
& Rows & Cols & Nonzeros & \#distinct & {\it maxval} & $|\Sigma|$ & \multicolumn{1}{c}{$S_B$} \\
\hline
{\tt airline78} & \num{14462943} & 29  & 72.66\%  & \num{7794}     & \num{226007}    & \num{29967}    & 4.5\%  \\
{\tt census}    & \num{2458285}  & 68  & 43.03\%  & \num{45}       & \num{3042}      & \num{331}      & 3.2\%  \\
{\tt covtype}   & \num{581012}   & 54  & 22.00\%  & \num{6682}     & \num{360779}    & \num{15952}    & 13.7\% \\
{\tt imagenet}  & \num{1262102}  & 900 & 30.99\%  & \num{824}      & \num{741448}    & \num{48772}    & 0.5\%  \\
{\tt mnist2m}   & \num{2000000}  & 784 & 25.25\%  & \num{255}      & \num{199915}    & \num{181260}   & 2.2\%  \\
\hdashline
{\tt glove}      & \num{400000} & 300 & 100.00\% & \num{717119}   & \num{215135700} & \num{56647729} & 99.7\% \\
{\tt higgs}     & \num{11000000} & 28  & 92.11\%  & \num{8083943}  & \num{226350398} & \num{8705556}  & 72.4\% \\
{\tt optical}   & \num{325834}   & 174 & 97.50\%  & \num{897176}   & \num{156108606} & \num{6929447}  & 67.7\% \\
{\pgsf}         & \num{2000000}  & 768 & 99.95\%  & \num{4744}     & \num{3642993}   & \num{2061458}  & 32.3\% \\
{\tt susy}      & \num{5000000}  & 18  & 98.82\%  & \num{20352142} & \num{366338544} & \num{23564797} & 95.0\% \\
\hline
\end{tabular}
\end{table}

We use the matrices summarized in Table~\ref{tab:datasets}.
The first group (above the dashed line) consists of the sparse matrices used in~\cite{Elgohary2018,FerraginaMGKNST22}.
The second group consists of matrices that are not sparse and whose CSRV alphabet is much larger than in the first group, with $|\Sigma|$ above two million against at most \num{181260}, since our techniques are specifically designed to improve \mmRepair on such inputs.
Besides \texttt{higgs}, \texttt{optical} and \texttt{susy}, taken from~\cite{FerraginaMGKNST22}, we consider the \texttt{glove.6B} word-vector matrix~\cite{pennington-etal-2014-glove} and the \pgsf matrix of hidden-state activations from layer 6 of 12 of GPT-2 run on a collection of Project Gutenberg science-fiction short stories.
Section~\ref{sec:quantization} additionally considers six quantized variants of \texttt{glove}, used to study the effect of the number of distinct values.

The last column of Table~\ref{tab:datasets} shows that the partitioning assigns to $S_B$ at most 13.7\% of the CSRV sequence on the sparse matrices, and between 32.3\% and 99.7\% on the dense ones.
The remapping criterion~\eqref{eq:map} was satisfied by all matrices except \texttt{glove} and \texttt{susy}, which have particularly large CSRV alphabets, so for these two matrices \our retains the original identifiers.


\subsection{Combining hybrid \mmRepair with column reordering}\label{sec:pc}

We combine hybrid \mmRepair with the column reordering of Section~\ref{sec:mmrepair}, in a variant called \ourPC.
The CSRV sequence is first partitioned into $S_A$ and $S_B$ as in \our, then the \texttt{PathCover} column reordering is applied to $S_A$ alone, and the pipeline continues exactly as in \our.
The reordering uses the setting that gave the best trade-off between compressed size and reordering time in~\cite{FerraginaMGKNST22}, \texttt{PathCover} on the similarity matrix pruned with $k=16$ and one permutation per row block.
Nonetheless, its cost is still very high, up to three orders of magnitude that of the compression itself, as Table~\ref{tab:compression-time} shows.
In exchange, \ourPC improves the compression ratio of \our on the sparse matrices, where column reordering helps, and matches it on the dense ones (Table~\ref{tab:compression-ratio} and Figure~\ref{fig:ratio-improvements}).

The two techniques are complementary: partitioning removes the entries whose digrams do not repeat often enough to pay for a grammar rule, whereas column reordering creates repeated digrams among the remaining entries, whose values already recur.
Applying the permutation to $S_A$ alone is thus natural, since the entries of $S_B$ take part in few repeated digrams and would contribute little to the similarity scores.
Partitioning is applied before the column reordering so that Definition~\ref{def:split} evaluates the digram frequencies of the original sequence, which a column permutation would change.

The column reordering does not interfere with the other steps of the pipeline.
Since it does not change the identifiers, the alphabet remapping gives the same output as in \our, and \textsc{ole} is executed after creating $C_A$.
The representation and the multiplication algorithms are therefore those of \our (Theorem~\ref{thm:hybrid}).



\subsection{Compression ratio}\label{sec:compression}

\begin{table}[t]
\centering
\caption{Compression ratios, expressed as percentages of the size of the original dense matrix.
Best value per row in bold.}
\label{tab:compression-ratio}
\small
\begin{tabular}{l|rr|rrrr}
\hline
\multirow{2}{*}{Matrix}
& \multicolumn{2}{c|}{Baseline}
& \multicolumn{4}{c}{\mmRepair} \\
\cline{2-7}
& \texttt{gzip} & \texttt{xz}
& \reans & \reansPC & \our & \ourPC \\
\hline
{\tt airline78} & 13.27\% & \textbf{7.14\%}  & 9.27\%  & 8.34\%  & 8.75\%  & 7.25\% \\
{\tt census}    & 5.54\%  & 2.74\%           & 1.53\%  & 1.30\%  & 1.50\%  & \textbf{1.27\%} \\
{\tt covtype}   & 6.25\%  & 3.39\%           & 3.87\%  & 3.26\%  & 3.38\%  & \textbf{2.88\%} \\
{\tt imagenet}  & 5.52\%  & 3.57\%           & 4.08\%  & 3.81\%  & 3.39\%  & \textbf{3.18\%} \\
{\tt mnist2m}   & 6.46\%  & \textbf{4.12\%}  & 5.33\%  & 5.31\%  & 4.48\%  & 4.48\% \\
\hdashline
{\tt glove}     & 59.93\% & \textbf{31.54\%} & 42.45\% & 42.46\% & 32.25\% & 32.25\% \\
{\tt higgs}     & 48.38\% & 30.01\%          & 38.05\% & 38.04\% & \textbf{28.64\%} & \textbf{28.64\%} \\
{\tt optical}   & 53.54\% & 24.88\%          & 34.31\% & 32.89\% & \textbf{23.68\%} & 24.21\% \\
{\pgsf}         & 25.49\% & \textbf{17.74\%} & 34.72\% & 34.72\% & 25.54\% & 25.55\% \\
{\tt susy}      & 53.27\% & \textbf{41.87\%} & 66.63\% & 66.57\% & 57.17\% & 57.17\% \\
\hline
\end{tabular}
\end{table}

The \textit{compression ratio} reported in Table~\ref{tab:compression-ratio} is the ratio between the size of the compressed representation and that of the original matrix stored as a plain array of 8-byte floating-point numbers, that is, $\textit{rows}\times\textit{cols}\times 8$ bytes.

\begin{figure}[t]
\centering
\includegraphics[width=\textwidth]{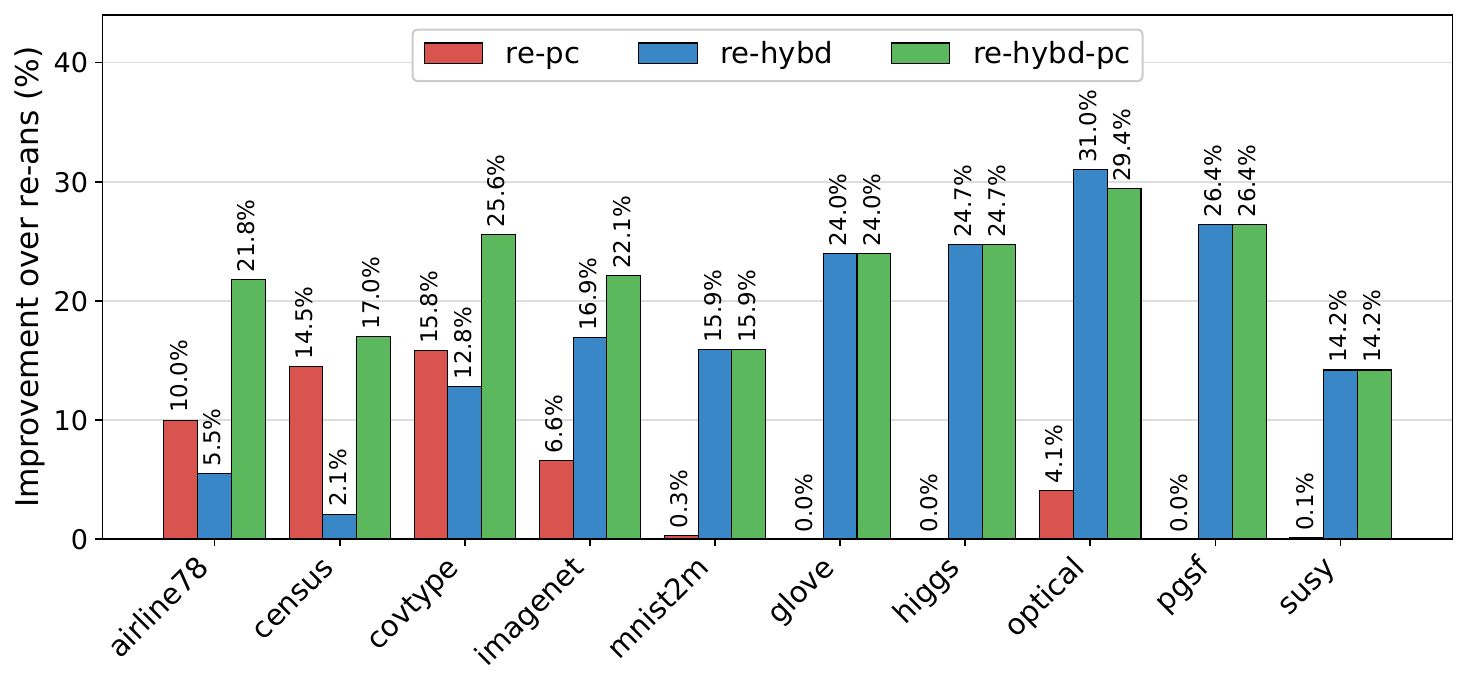}
\caption{Relative improvement in compression ratio of \reansPC, \our and \ourPC over \reans, computed as $(c_o-c_r)/c_o$, where $c_o$ and $c_r$ are the compression ratios of \reans and of the compared method.}
\label{fig:ratio-improvements}
\end{figure}

Table~\ref{tab:compression-ratio} shows that \our achieves a better compression ratio than \reans on all datasets, and Figure~\ref{fig:ratio-improvements} reports the corresponding relative improvements.
Among \reans, \reansPC and \our, \reansPC is slightly better on \texttt{airline78}, \texttt{census}, and \texttt{covtype}, where it improves over \reans by 10.0\%, 14.5\% and 15.8\%, while \our is the best on the remaining seven, often by a wide margin.
The improvements of \our are particularly pronounced for matrices that are less sparse and whose CSRV representation has a large alphabet: the ratio drops from 42.45\% to 32.25\% on \texttt{glove}, from 38.05\% to 28.64\% on \texttt{higgs}, from 34.31\% to 23.68\% on \texttt{optical}, from 66.63\% to 57.17\% on \texttt{susy}, and from 34.72\% to 25.54\% on \pgsf, that is, relative improvements between 14.2\% and 31.0\%.
On the same matrices, \reansPC improves over \reans by at most 4.1\% (\texttt{optical}): column reordering cannot create repetitions where the alphabet is too large for digrams to repeat, which is exactly where the partitioning helps.
These are also the matrices with the largest fraction of the CSRV sequence assigned to $S_B$ (Table~\ref{tab:datasets}), which confirms that the entries routed to the entropy coder are precisely those on which \Repair was spending most of its space.
On the sparse matrices of the first group $S_B$ holds at most 13.7\% of the entries, so the partitioning contributes little, and the gains, from 2.1\% on \texttt{census} to 16.9\% on \texttt{imagenet}, come mostly from the row ordering and the remapping (Section~\ref{sec:components}).
The gain on \texttt{susy} is worth noting, since this matrix, with more than 98\% of nonzero entries and over 20 million distinct values, is cited in~\cite{Navarro26survey} as the kind of input on which no computation-friendly format helps and the plain representation is the right choice, whereas \our brings its compressed size from 66.63\% to 57.17\% while retaining multiplication over the compressed data.

Combining the two techniques in \ourPC gives the best ratio of all variants on the sparse matrices on which column reordering helps,
that is, 
between 17\% and 26\% less than \reans (Figure~\ref{fig:ratio-improvements}).
On the remaining matrices \ourPC is comparable to \our, since column reordering brings little to \reans on them as well.
Its ratio is identical on \texttt{glove}, \texttt{higgs}, \texttt{mnist2m} and \texttt{susy}, and within 0.6 points on \texttt{optical} (24.21\% against 23.68\%) and \pgsf (25.55\% against 25.54\%).
Reordering therefore pays off only where the grammar compresses most of the sequence.

Compared with \texttt{xz}, which does not support multiplication over the compressed data, \our is more compact on \texttt{census}, \texttt{higgs}, \texttt{imagenet}, and \texttt{optical}, comparable on \texttt{covtype} and \texttt{glove}, and larger on the others.
\ourPC is in addition more compact than \texttt{xz} on \texttt{covtype} and comparable to it on \texttt{airline78}.
Compared with \texttt{gzip}, \our is clearly more compact on eight of the ten matrices, comparable on \pgsf, 
and larger only on \texttt{susy}.


\subsection{Compression time}\label{sec:compression-time}

\begin{table*}[t]
\centering
\setlength{\tabcolsep}{6pt}
\caption{Compression times in seconds.}
\label{tab:compression-time}
\small
\begin{tabular}{l|rr|rrrr}
\hline
\multirow{2}{*}{Matrix}
& \multicolumn{2}{c|}{Baseline}
& \multicolumn{4}{c}{\mmRepair} \\
\cline{2-7}
& \texttt{gzip} & \texttt{xz}
& \reans & \reansPC & \our & \ourPC \\
\hline
{\tt airline78} & 92.10  & \num{1852.06}  & 375.81  & \num{4340.11}   & \num{1062.10}  & \num{4874.00} \\
{\tt census}    & 25.36  & 290.50         & 40.30   & \num{3403.37}   & 86.00          & \num{3496.71} \\
{\tt covtype}   & 3.07   & 58.71          & 5.93    & 455.31          & 52.47          & 473.89 \\
{\tt imagenet}  & 180.83 & \num{2062.24}  & 562.73  & \num{283583.64} & \num{1267.62}  & \num{280717.79} \\
{\tt mnist2m}   & 156.60 & \num{3154.99}  & 401.40  & \num{341737.83} & \num{2601.48}  & \num{339112.74} \\
\hdashline
{\tt glove}     & 30.70  & \num{1036.64}  & 165.70  & \num{10391.59}  & \num{683.65}  & \num{9170.82} \\
{\tt higgs}     & 207.53 & \num{2426.91}  & 330.75  & \num{3713.33}   & \num{3007.10}  & \num{4648.80} \\
{\tt optical}   & 14.43  & 287.30         & 50.34   & \num{2926.86}   & 282.94         & \num{3018.83} \\
{\pgsf}         & 875.12 & \num{12988.75} & \num{2129.43} & \num{377972.82} & \num{13765.55} & \num{360055.71} \\
{\tt susy}      & 70.66  & 746.80         & 102.76  & 884.91          & 406.09         & 893.86 \\
\hline
\end{tabular}
\end{table*}

Table~\ref{tab:compression-time} reports the compression times.
Among the general-purpose compressors, \texttt{gzip} is between $1.5\times$ and $5.4\times$ faster than \reans, whereas \texttt{xz} is always slower than \reans.
The column reordering of \reansPC is very expensive, even though \texttt{PathCover} is among the fastest of the heuristics evaluated in~\cite{FerraginaMGKNST22}: on \texttt{mnist2m} and \texttt{imagenet}, with 784 and 900 columns, it takes several days, with slowdowns over \reans of $851\times$ and above $500\times$.

In contrast, the compression time of \our ranges from $2.1\times$ (\texttt{census}) to $9.1\times$ (\texttt{higgs}) that of \reans, and is comparable to that of \texttt{xz}, being lower on eight of the ten matrices and within $1.3\times$ on the others.
The overhead over \reans comes from the computation of the digram frequencies, from the alphabet remapping, and from the additional encoding of $S_B$ and $\mathcal{C}_A$.
Nonetheless, \our provides substantial compression improvements with a much smaller compression-time overhead than \reansPC.

The compression time of \ourPC is close to that of \reansPC, between $0.88\times$ (\texttt{glove}) and $1.25\times$ (\texttt{higgs}) of the latter, and between $8.7\times$ (\texttt{susy}) and $845\times$ (\texttt{mnist2m}) that of \reans.
At first sight one could expect \ourPC to be clearly faster, since it computes the permutation on $S_A$ alone, which on \texttt{glove} holds only 0.3\% of the entries.
It is not, because the cost of building the column-similarity matrix depends on the number of rows and on the square of the number of columns rather than on the number of nonzeros that remain.
The reordering is indeed slightly cheaper on the matrices with the most columns (\texttt{imagenet}, \texttt{mnist2m} and \pgsf) and on \texttt{glove}, by 1\% to 12\%, but on the other six matrices this saving is smaller than the cost of the partitioning and of the encoding of $S_B$, and \ourPC ends up slower than \reansPC, by up to 25\% on \texttt{higgs}, whose 28 columns make the reordering itself cheap.
Since \ourPC improves compression only on the sparse matrices (Section~\ref{sec:compression}), the choice between \our and \ourPC is a trade-off between compression time and space on those inputs.


\subsection{Component analysis}\label{sec:components}

\begin{figure}[t]
\centering
\includegraphics[width=\textwidth]{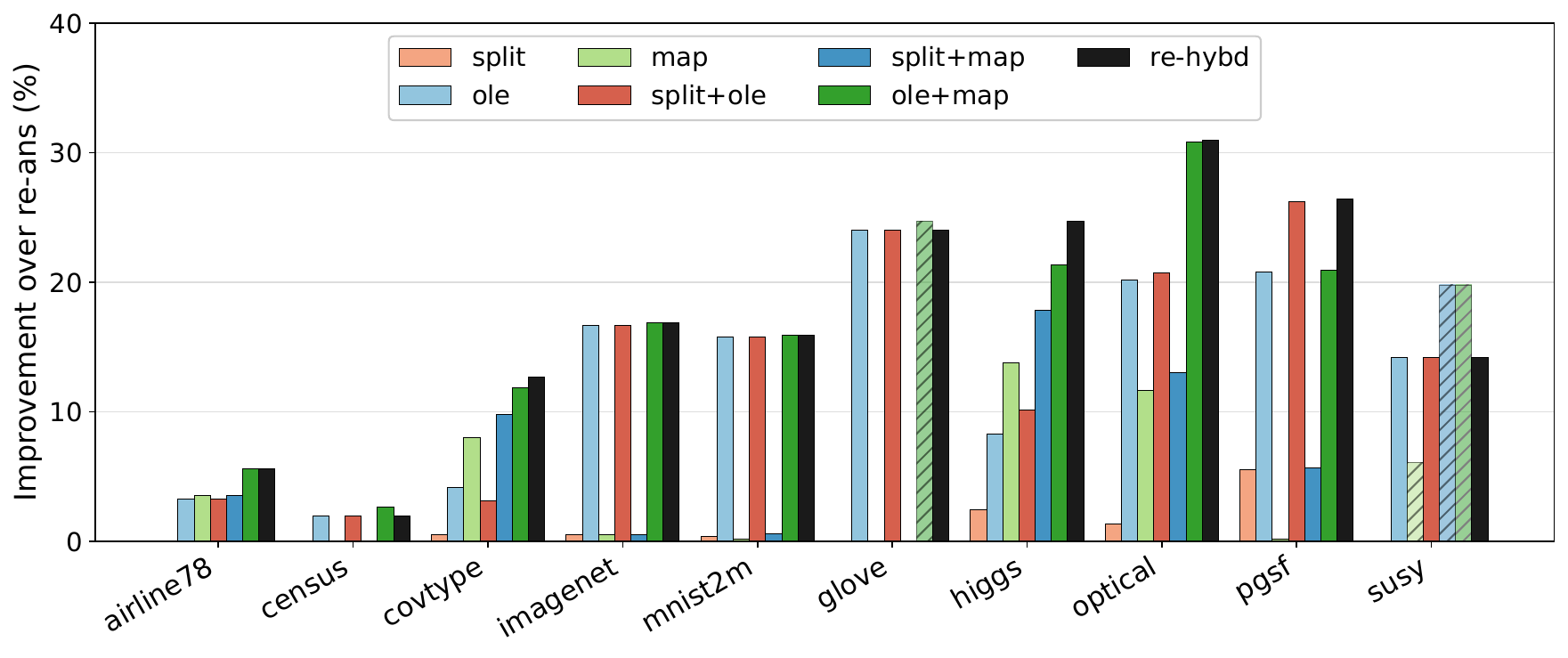}
\caption{Relative improvement in compression ratio over \reans of each configuration of Table~\ref{tab:components}, computed as in Figure~\ref{fig:ratio-improvements}.
Hatched bars are the configurations in which the remapping is forced against criterion~\eqref{eq:map}.}
\label{fig:components}
\end{figure}

To isolate the contribution of each technique introduced in \our, we evaluate the partitioning (\textsc{split}), the row ordering with delta encoding (\textsc{ole}) and the alphabet remapping (\textsc{map}) separately and in combination.
Table~\ref{tab:components} reports the resulting compression ratios, with \reans as the baseline.
Figure~\ref{fig:components} shows the corresponding relative improvements over \reans.
On \texttt{glove} and \texttt{susy}, where criterion~\eqref{eq:map} rejects the remapping, it was forced: these configurations save space at a memory cost during multiplication that \our does not pay.

Among the individual techniques, \textsc{ole} provides the largest and most consistent improvement.
It reduces the compression ratio on all matrices, by 14\% to 24\% on \texttt{glove}, \texttt{optical}, \pgsf, \texttt{imagenet}, \texttt{mnist2m} and \texttt{susy}, and by 2\% to 8\% on the others.
Since the gaps of a sorted row add up to its largest identifier, the gain grows with the row length, which is why the five matrices with the most columns (\texttt{imagenet}, \texttt{mnist2m}, \pgsf, \texttt{glove} and \texttt{optical}) are among the six, and with the number of terminals left in $\mathcal{C}$ and the size of their identifiers, which is why \texttt{susy}, with only 18 columns but 95\% of its entries left as terminals (the $S_B$ column of Table~\ref{tab:datasets}) and the largest identifiers of all, is there as well.

Alphabet remapping alone is beneficial on several matrices, most notably \texttt{higgs} and \texttt{optical}, has almost no effect on \texttt{census} and \texttt{mnist2m}, and slightly increases the ratio on \texttt{glove}, consistently with criterion~\eqref{eq:map}, which rejects it there.
The partitioning alone has an even smaller effect, with negligible changes on \texttt{airline78}, \texttt{census}, \texttt{glove}, and \texttt{susy}.
The reason is that moving an entry from $\mathcal{C}$ to $S_B$ does not by itself make it cheaper: both sequences are encoded with \ansfold, and the gain of the partitioning comes from the smaller grammar and from the cleaner $S_A$ only when $S_B$ is itself encoded well.

\begin{table}[t]
\centering
\setlength{\tabcolsep}{2.5pt}
\caption{Compression ratio (\%) of each technique.
\textsc{split} denotes the hybrid partitioning into $S_A$ and $S_B$, \textsc{ole} the row ordering and delta encoding, and \textsc{map} the alphabet remapping.
\our is the combination of all three.
Values marked $^\dagger$ were obtained by forcing the remapping on \texttt{glove} and \texttt{susy}, where criterion~\eqref{eq:map} rejects it, so they are not configurations that \our uses.
Best value per row in bold.
}
\label{tab:components}
\small
\begin{tabular}{l|r|rrr|rrr|r}
\hline
\multirow{2}{*}{Matrix} & \multirow{2}{*}{\reans}
& \multicolumn{3}{c|}{One technique}
& \multicolumn{3}{c|}{Two techniques}
& \multirow{2}{*}{\our} \\
\cline{3-8}
& & \textsc{split} & \textsc{ole} & \textsc{map}
& \textsc{split+ole} & \textsc{split+map} & \textsc{ole+map} & \\
\hline
{\tt airline78} & 9.27\%  & 9.27\%  & 8.97\%  & 8.94\%  & 8.97\%  & 8.94\%  & \textbf{8.75\%}  & \textbf{8.75\%} \\
{\tt census}    & 1.53\%  & 1.53\%  & 1.50\%  & 1.53\%  & 1.50\%  & 1.53\%  & \textbf{1.49\%}  & 1.50\% \\
{\tt covtype}   & 3.87\%  & 3.85\%  & 3.71\%  & 3.56\%  & 3.75\%  & 3.49\%  & 3.41\%           & \textbf{3.38\%} \\
{\tt imagenet}  & 4.08\%  & 4.06\%  & 3.40\%  & 4.06\%  & 3.40\%  & 4.06\%  & \textbf{3.39\%}  & \textbf{3.39\%} \\
{\tt mnist2m}   & 5.33\%  & 5.31\%  & 4.49\%  & 5.32\%  & 4.49\%  & 5.30\%  & \textbf{4.48\%}  & \textbf{4.48\%} \\
\hdashline
{\tt glove}      & 42.45\% & 42.45\% & \textbf{32.25\%} & 42.64\%$^\dagger$ & \textbf{32.25\%} & 42.64\%$^\dagger$ & 31.97\%$^\dagger$ & \textbf{32.25\%} \\
{\tt higgs}     & 38.05\% & 37.13\% & 34.91\% & 32.81\% & 34.20\% & 31.26\% & 29.94\%          & \textbf{28.64\%} \\
{\tt optical}   & 34.31\% & 33.84\% & 27.38\% & 30.32\% & 27.19\% & 29.85\% & 23.73\%          & \textbf{23.68\%} \\
{\pgsf}         & 34.72\% & 32.81\% & 27.50\% & 34.67\% & 25.62\% & 32.75\% & 27.44\%          & \textbf{25.54\%} \\
{\tt susy}      & 66.63\% & 66.63\% & \textbf{57.17\%} & 62.59\%$^\dagger$ & \textbf{57.17\%} & 53.44\%$^\dagger$ & 53.44\%$^\dagger$ & \textbf{57.17\%} \\
\hline
\end{tabular}
\end{table}

The combination of \textsc{ole} and \textsc{map} improves substantially over either technique alone, for example from 38.05\% to 29.94\% on \texttt{higgs} and from 34.31\% to 23.73\% on \texttt{optical}.
Adding the partitioning improves compression further on several matrices, most notably \pgsf, from 27.44\% with \textsc{ole+map} to 25.54\%, and \texttt{higgs}, from 29.94\% to 28.64\%.
The effect of each technique is therefore dataset-dependent, but their combination consistently improves over \reans: \our obtains the lowest ratio, or a value tied for the lowest, on nine of the ten matrices, the exception being \texttt{census}, where \textsc{ole+map} is marginally better (1.49\% against 1.50\%).
Forcing the remapping on \texttt{glove} and \texttt{susy} would lower their ratios further, to 31.97\% and 53.44\%, at the cost of keeping a reverse mapping of 226.6\,MB and 94.3\,MB in memory during multiplication, which criterion~\eqref{eq:map} avoids.
Beyond compression, the partitioning is the technique that reduces the grammar, and hence the auxiliary array $W$ used during multiplication, a benefit on peak memory that we discuss next.

\subsection{Matrix--vector multiplication: peak memory}\label{sec:multiplication}


Following the experiments in~\cite{FerraginaMGKNST22}, we perform $500$ alternating right and left multiplications according to \[ \mathbf{y}_i = M\mathbf{x}_i,\qquad \mathbf{z}_i = \mathbf{y}_i^{T}M,\qquad \mathbf{x}_{i+1} = \mathbf{z}_i/\|\mathbf{z}_i\|_\infty , \] where $\|\mathbf{z}_i\|_\infty$ is the maximum absolute value among the entries of $\mathbf{z}_i$.
We report here the peak memory of the whole process, which includes the compressed representations of $\mathcal{R}_A$, $\mathcal{C}_A$, $S_B$, the array $V$, the auxiliary array $W$, the decoded \wcode when the remapping is applied, and the vectors.

Table~\ref{tab:peak-memory}
shows that \our consistently reduces peak memory with respect to \reans.
The largest reductions are on \pgsf, \texttt{higgs} and \texttt{glove}, by 26.1\%, 25.3\% and 23.9\%, followed by \texttt{optical} (17.1\%) and \texttt{covtype} (14.7\%), while the remaining matrices show reductions between 3.6\% on \texttt{airline78} and 13.7\% on \texttt{imagenet}.
Since the compressed representation is held in memory during multiplication, peak memory closely tracks the compression ratio: on \texttt{glove}, \texttt{higgs}, \pgsf and \texttt{susy} the reductions in peak memory are within one point of the reductions in compressed size.
The reduction on \texttt{susy} is smaller than on the other dense matrices because its 20.4M distinct values make the uncompressed array $V$ alone 22.6\% of the dense matrix, a cost that every representation pays.
\reansPC, in contrast, reduces peak memory by only 6.3\% to 11.7\% on \texttt{airline78}, \texttt{census}, \texttt{covtype} and \texttt{imagenet}, and essentially not at all on the other matrices, where it even increases it slightly on \pgsf and \texttt{susy}.

\ourPC has the lowest peak memory on \texttt{airline78}, \texttt{census}, \texttt{covtype} and \texttt{imagenet}, with reductions between 12.4\% and 21.8\% over \reans.
On these sparse matrices the grammar holds most of the sequence, so the smaller grammar produced by the column reordering reduces both the compressed representation and the array $W$, and the two techniques add up: the reductions of \ourPC are close to the sum of those of \reansPC and \our, for example 21.8\% against 11.7\% and 3.6\% on \texttt{airline78}.
On the other matrices it behaves like \our, with identical values on \texttt{mnist2m}, \texttt{higgs}, \texttt{glove} and \texttt{susy}, and values within 0.5 points on \texttt{optical} and \pgsf, since there the grammar is small and reordering has little to act on.

{
To isolate the effect of the partitioning on memory, we also ran the multiplications with the \textsc{ole+map} configuration of Table~\ref{tab:components}, that is, \our without \textsc{split}.
Its peak memory is within 0.1 points of that of \our in two kinds of matrices: on \texttt{airline78}, \texttt{census}, \texttt{imagenet} and \texttt{mnist2m}, where the partitioning moves at most 4.5\% of the entries to $S_B$ (Table~\ref{tab:datasets}) and leaves the grammar almost unchanged, and on \texttt{glove} and \texttt{susy}, where it moves almost all of them but the grammar is already small, since few digrams repeat.
It is higher where the partitioning removes many rules from a large grammar: 40.13\% against 35.19\% on \texttt{higgs}, 37.26\% against 32.87\% on \pgsf, 33.29\% against 32.37\% on \texttt{optical} and 8.98\% against 8.56\% on \texttt{covtype}.
The partitioning is thus the technique that reduces the array $W$, and with it the peak memory of multiplication.
}

\begin{table}[t]
\centering
\setlength{\tabcolsep}{7pt}
\caption{Peak memory during matrix--vector multiplication, as a percentage of the size of the original dense matrix, and relative reduction with respect to \reans, computed as $(p_o-p_r)/p_o$ where $p_o$ and $p_r$ are the peak memory of \reans and of the compared method.
Best value per row in bold.}
\label{tab:peak-memory}
\small
\begin{tabular}{l|rrrr|rrr}
\hline
\multirow{2}{*}{Matrix}
& \multicolumn{4}{c|}{Peak memory}
& \multicolumn{3}{c}{Reduction w.r.t.\ \reans} \\
\cline{2-8}
& \reans & \reansPC & \our & \ourPC & \reansPC & \our & \ourPC \\
\hline
{\tt airline78} & 15.39\% & 13.59\% & 14.83\% & \textbf{12.03\%} & 11.7\% & 3.6\%  & 21.8\% \\
{\tt census}    & 4.12\%  & 3.76\%  & 3.96\%  & \textbf{3.61\%}  & 8.6\%  & 3.9\%  & 12.4\% \\
{\tt covtype}   & 10.03\% & 8.98\%  & 8.56\%  & \textbf{7.94\%}  & 10.4\% & 14.7\% & 20.8\% \\
{\tt imagenet}  & 5.20\%  & 4.88\%  & 4.49\%  & \textbf{4.22\%}  & 6.3\%  & 13.7\% & 18.8\% \\
{\tt mnist2m}   & 6.85\%  & 6.83\%  & \textbf{5.96\%}  & \textbf{5.96\%}  & 0.2\%  & 13.0\% & 13.0\% \\
\hdashline
{\tt glove}     & 43.64\% & 43.64\% & \textbf{33.20\%} & \textbf{33.20\%} & 0.0\%  & 23.9\% & 23.9\% \\
{\tt higgs}     & 47.09\% & 47.00\% & \textbf{35.19\%} & \textbf{35.19\%} & 0.2\%  & 25.3\% & 25.3\% \\
{\tt optical}   & 39.07\% & 37.68\% & \textbf{32.37\%} & 32.83\% & 3.5\%  & 17.1\% & 16.0\% \\
{\pgsf}         & 44.48\% & 44.50\% & \textbf{32.87\%} & 32.90\% & -0.1\% & 26.1\% & 26.0\% \\
{\tt susy}      & 73.33\% & 73.55\% & \textbf{63.64\%} & \textbf{63.64\%} & -0.3\% & 13.2\% & 13.2\% \\
\hline
\end{tabular}
\end{table}


\subsection{Matrix--vector multiplication: running time}\label{sec:running-time}

Table~\ref{tab:running-time} reports the average running time per iteration of the $500$ multiplications as in~\cite{FerraginaMGKNST22}.
The additional work of \our, namely delta decoding each row and scanning $S_B$ symbol by symbol, makes multiplication slower in most cases: the slowdown is negligible on \texttt{imagenet} and \texttt{mnist2m}, and moderate elsewhere, between $0.67\times$ on \texttt{covtype} and $0.87\times$ on \texttt{glove}.
It is not proportional to the size of $S_B$.
On \texttt{glove}, where $S_B$ holds 99.7\% of the entries and the multiplication is essentially a sparse product over an \ansfold-encoded sequence, \our runs at $0.87\times$, whereas on \texttt{covtype} and \texttt{census}, whose compressed representations are tiny, the per-row delta decoding of $\mathcal{C}_A$ and $S_B$ and the duplicated delimiters (the $O(n)$ term of Theorem~\ref{thm:hybrid}) weigh more than the entropy decoding and bring the speedup down to $0.67\times$ and $0.71\times$.
The exception is \pgsf, where \our is $1.27\times$ faster than \reans and $1.16\times$ faster than \reansPC, because its much smaller grammar reduces the cost of evaluating the array $W$, as discussed after Theorem~\ref{thm:hybrid}.
\reansPC is faster than \reans on eight of the ten matrices, between $1.07\times$ and $1.26\times$, since the reordered columns yield a smaller grammar, and it is the fastest method on \texttt{census}, \texttt{covtype}, \texttt{higgs} and \texttt{optical}.

\ourPC combines the two effects on the sparse matrices: it is the fastest method on \texttt{airline78}, \texttt{imagenet} and \texttt{mnist2m}, at $1.15\times$, $1.22\times$ and $1.12\times$ the speed of \reans, and on \texttt{census} and \texttt{covtype} it is faster than \our but still slower than \reans ($0.85\times$ and $0.75\times$).
On the dense matrices it is slower than \reans, like \our, between $0.53\times$ on \texttt{optical} and $0.91\times$ on \texttt{glove}, and on \pgsf it is $1.20\times$ faster than \reans but slower than \our.

Overall, \our achieves lower peak memory on every input, with a speedup over \reans between $0.7\times$ and $1.3\times$, and the trade-off is most favorable on the dense matrices with many distinct values that motivated this work.

\begin{table}[t]
\centering
\setlength{\tabcolsep}{7pt}
\caption{Average running time per iteration of matrix--vector multiplication, in seconds, and speedup with respect to \reans, where speedups below $1\times$ indicate a slowdown.
Best value per row in bold.}
\label{tab:running-time}
\small
\begin{tabular}{l|rrrr|rrr}
\hline
\multirow{2}{*}{Matrix}
& \multicolumn{4}{c|}{Running time (s)}
& \multicolumn{3}{c}{Speedup w.r.t.\ \reans} \\
\cline{2-8}
& \reans & \reansPC & \our & \ourPC & \reansPC & \our & \ourPC \\
\hline
{\tt airline78} & 2.84  & 2.54          & 3.56           & \textbf{2.46} & 1.12$\times$ & 0.80$\times$ & 1.15$\times$ \\
{\tt census}    & 0.17  & \textbf{0.15} & 0.24           & 0.20          & 1.13$\times$ & 0.71$\times$ & 0.85$\times$ \\
{\tt covtype}   & 0.06  & \textbf{0.05} & 0.09           & 0.08          & 1.20$\times$ & 0.67$\times$ & 0.75$\times$ \\
{\tt imagenet}  & 3.18  & 2.86          & 3.29           & \textbf{2.60} & 1.11$\times$ & 0.97$\times$ & 1.22$\times$ \\
{\tt mnist2m}   & 6.40  & 6.00          & 6.28           & \textbf{5.71} & 1.07$\times$ & 1.02$\times$ & 1.12$\times$ \\
\hdashline
{\tt glove}     & \textbf{2.06} & 2.22  & 2.38           & 2.27          & 0.93$\times$ & 0.87$\times$ & 0.91$\times$ \\
{\tt higgs}     & 8.05  & \textbf{7.43} & 11.00          & 11.65         & 1.08$\times$ & 0.73$\times$ & 0.69$\times$ \\
{\tt optical}   & 1.11  & \textbf{0.88} & 1.44           & 2.09          & 1.26$\times$ & 0.77$\times$ & 0.53$\times$ \\
{\pgsf}         & 39.69 & 36.20         & \textbf{31.16} & 32.97         & 1.10$\times$ & 1.27$\times$ & 1.20$\times$ \\
{\tt susy}      & \textbf{2.90} & 2.93  & 4.16           & 3.46          & 0.99$\times$ & 0.70$\times$ & 0.84$\times$ \\
\hline
\end{tabular}
\end{table}

\subsection{Effect of the number of distinct values: quantized GloVe}\label{sec:quantization}

\begin{figure}[t]
\centering
\includegraphics[width=0.95\textwidth]{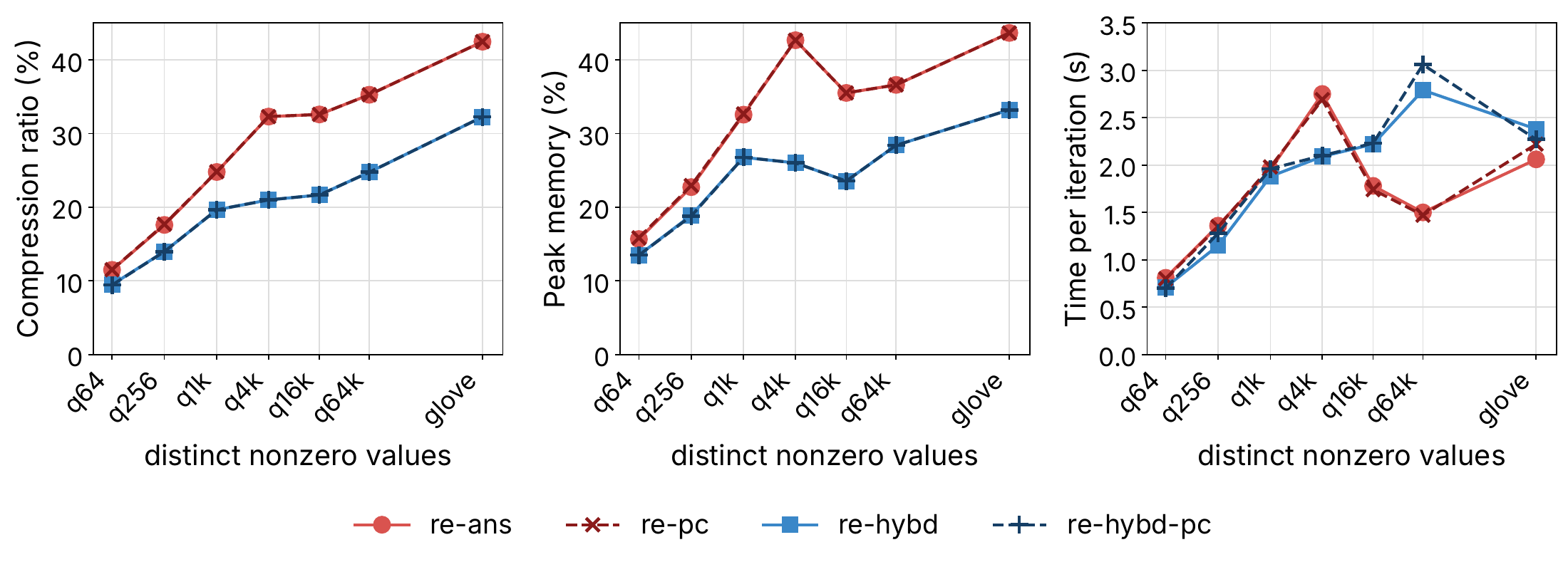}
\caption{Compression ratio, peak memory and running time of \reans, \reansPC, \our and \ourPC on \texttt{glove} as a function of the number of distinct nonzero values (logarithmic scale), from the 64 values of \texttt{glove.q64} to the \num{717119} values of the original matrix.
The column-reordering variants are drawn with dashed lines and cross markers, so that they remain visible where they coincide with \reans and \our.}
\label{fig:quantization}
\end{figure}

The matrices on which \our gains most are those with a large CSRV alphabet, which grows with the number of distinct nonzero values.
To isolate this effect, we quantize the \texttt{glove} matrix, which is fully dense and has \num{717119} distinct values, to $K=64$, $256$, $2^{10}$, $2^{12}$, $2^{14}$ and $2^{16}$ levels, obtaining \texttt{glove.q64}, \texttt{glove.q256}, \texttt{glove.q1k}, \texttt{glove.q4k}, \texttt{glove.q16k} and \texttt{glove.q64k}.
The quantization is uniform and global: with $v_{\min}$ and $v_{\max}$ the smallest and largest entries of the whole matrix and $\Delta=(v_{\max}-v_{\min})/(K-1)$, every entry $x$ is replaced by its nearest level, $v_{\min}+\mathrm{round}\big((x-v_{\min})/\Delta\big)\Delta$, with the same range for all columns.
Since the values are concentrated around zero, the levels in the tails remain empty, and for $K\geq 256$ the number of distinct values is smaller than $K$.
Table~\ref{tab:quantization-datasets} lists the seven matrices evaluated.
Criterion~\eqref{eq:map} selects the remapping on the six quantized variants and rejects it only on \texttt{glove}.
Figure~\ref{fig:quantization} reports the compression ratios, the peak memory and the running times.

As expected, all methods compress less as the number of distinct values grows.
The improvement of \our over \reans is present at every level: 17.4\% on \texttt{glove.q64}, 20.5\% on \texttt{glove.q256}, 20.7\% on \texttt{glove.q1k}, 35.0\% on \texttt{glove.q4k}, 33.5\% on \texttt{glove.q16k}, 29.8\% on \texttt{glove.q64k} and 24.0\% at full precision.
The $S_B$ column of Table~\ref{tab:quantization-datasets} explains the difference.
Up to \num{1024} levels $S_B$ holds at most 4.9\% of the entries, so the gain comes mainly from the row ordering and the remapping.
On \texttt{glove.q4k} it jumps to 74.1\%, and from \texttt{glove.q16k} on it exceeds 99\%, so the grammar component is almost empty and the gain is that of the entropy coder, helped by the remapping on the quantized variants but not at full precision.

\begin{table*}[t]
\centering
\setlength{\tabcolsep}{5pt}
\caption{\texttt{glove} and its quantized variants, sorted by number of distinct nonzero values, with the columns of Table~\ref{tab:datasets}.
``Levels'' is the number of quantization levels.
The last column reports the fraction of $S$ assigned to $S_B$ by \our, with a single block.}
\label{tab:quantization-datasets}
\small
 \resizebox{\textwidth}{!}{%
\begin{tabular}{l|r|rr|rr|rr|r}
\hline
\multirow{2}{*}{Matrix} & \multirow{2}{*}{Levels}
& \multicolumn{2}{c|}{Dimensions}
& \multicolumn{2}{c|}{Density}
& \multicolumn{2}{c|}{CSRV} & Partition \\
\cline{3-9}
& & Rows & Cols & Nonzeros & \#distinct & {\it maxval} & $|\Sigma|$ & \multicolumn{1}{c}{$S_B$} \\
\hline
{\tt glove.q64}  & 64          & \num{400000} & 300 & 100.00\% & \num{64}     & \num{18613}     & \num{12590}    & 0.4\%  \\
{\tt glove.q256} & 256         & \num{400000} & 300 & 100.00\% & \num{239}    & \num{71413}     & \num{45853}    & 0.6\%  \\
{\tt glove.q1k}  & \num{1024}  & \num{400000} & 300 & 100.00\% & \num{892}    & \num{267420}    & \num{164789}   & 4.9\%  \\
{\tt glove.q4k}  & \num{4096}  & \num{400000} & 300 & 100.00\% & \num{3294}   & \num{988031}    & \num{583634}   & 74.1\% \\
{\tt glove.q16k} & \num{16384} & \num{400000} & 300 & 100.00\% & \num{11840}  & \num{3551977}   & \num{2025863}  & 99.5\% \\
{\tt glove.q64k} & \num{65536} & \num{400000} & 300 & 100.00\% & \num{41525}  & \num{12457310}  & \num{6806213}  & 99.7\% \\
{\tt glove}      & --          & \num{400000} & 300 & 100.00\% & \num{717119} & \num{215135700} & \num{56647729} & 99.7\% \\
\hline
\end{tabular}
}
\end{table*}

The peak memory of \reans is not monotone: it reaches 42.65\% on \texttt{glove.q4k}, above the 35.50\% of \texttt{glove.q16k} and the 36.59\% of \texttt{glove.q64k}, and close to the 43.64\% at full precision, and its running time also peaks there, at 2.75\,s.
The reason is the grammar: with a few thousand distinct values \Repair still replaces many digrams that occur two or three times, so the grammar of \texttt{glove.q4k} is large although it saves little space, and both the array $W$ and the time to evaluate it grow with the number of rules.
With more distinct values fewer digrams repeat, the grammar shrinks and the terminals move to $\mathcal{C}$, where they cost less.
\our removes the rules that do not pay, and the peak disappears: from \texttt{glove.q1k} to \texttt{glove.q64k} its peak memory stays between 23.54\% and 28.40\%, against 32.55\% to 42.65\% for \reans.
Its reduction over \reans is the largest on \texttt{glove.q4k}, 39.0\%, where it is also $1.32\times$ faster, and its peak memory is the lowest of this range on \texttt{glove.q16k}, where $S_B$ holds more than 99\% of the entries and the array $W$ is almost empty.

The running times show both behaviors observed in Section~\ref{sec:running-time}.
Up to \num{4096} levels \our is $1.04\times$ to $1.32\times$ faster than \reans, because of its smaller grammar, while it runs at $0.80\times$ on \texttt{glove.q16k}, $0.54\times$ on \texttt{glove.q64k} and $0.87\times$ at full precision.
We attribute the slowdown on the two largest quantized variants to the remapping, since every terminal is resolved by a random access to the decoded \wcode, which has 2.0 million entries on \texttt{glove.q16k} and 6.8 million on \texttt{glove.q64k}, whereas at full precision the pair $\langle\ell,j\rangle$ is computed from the identifier itself by one division, with no table lookup.

Finally, column reordering shows no improvement at any quantization level.
\reansPC is within 0.1 points of \reans in compression ratio and 0.3 points in peak memory at every level, and within about 2\% in running time on the quantized variants, 8\% slower at full precision.
\ourPC is within 0.2 points of \our in peak memory and never worse in compression ratio, by at most 0.05 points better on \texttt{glove.q64} and \texttt{glove.q256}, and its running time is within 5\% of that of \our except on \texttt{glove.q256} and \texttt{glove.q64k}, where it is 11\% and 10\% slower, the cost of the reordered $\mathcal{C}_A$ rows without any gain in grammar size.


\begin{figure}[t]
\centering
\includegraphics[width=0.975\linewidth]{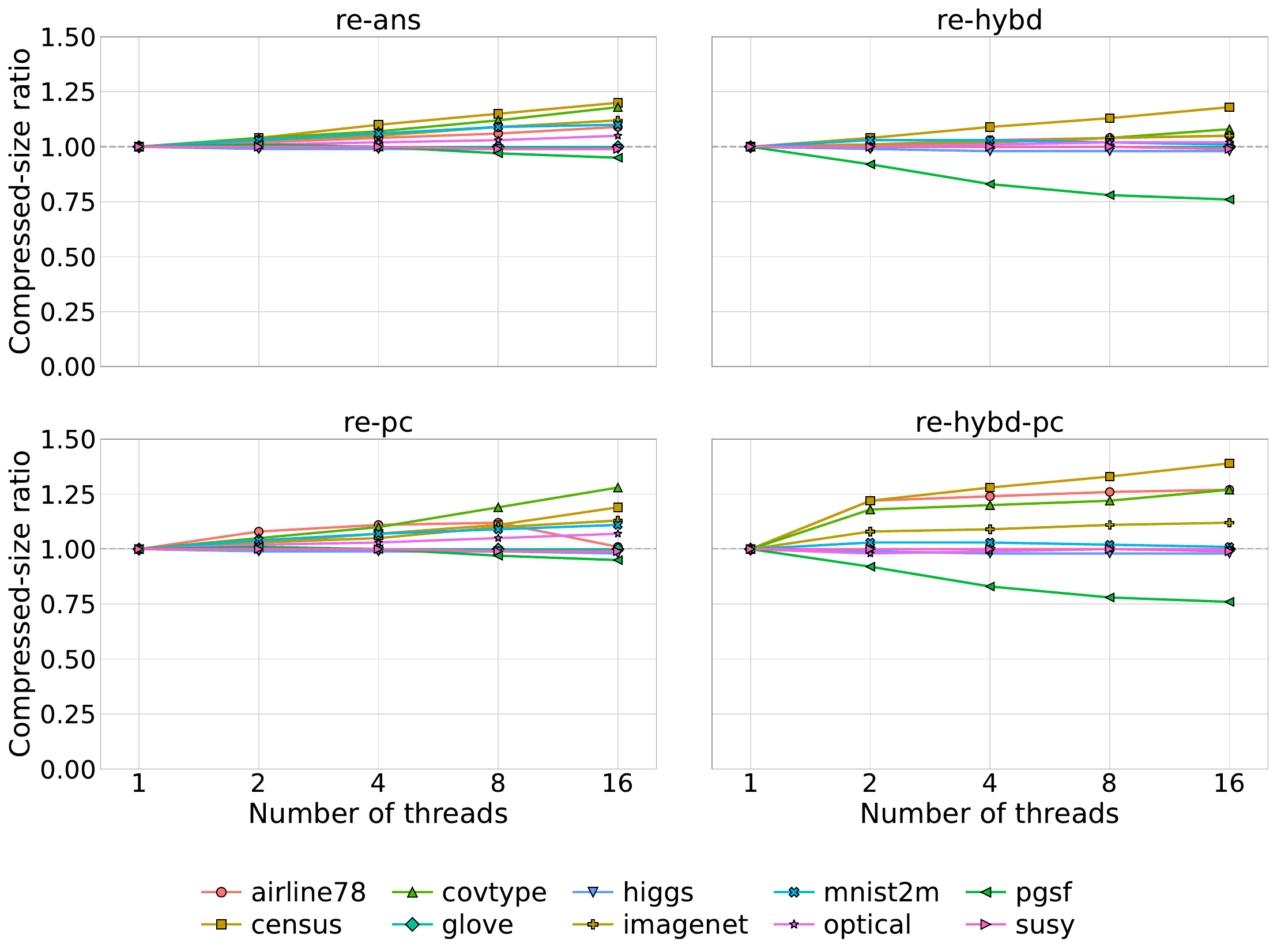}
\caption{Compressed size of the matrices of Table~\ref{tab:datasets} when the input is divided into $p$ row blocks, relative to the single-threaded representation, for \reans, \reansPC, \our and \ourPC.}
\label{fig:parallel-compression}
\end{figure}

\subsection{Parallel scalability}\label{sec:parallel-exp}

\begin{figure}[t]
\centering
\includegraphics[width=0.975\linewidth]{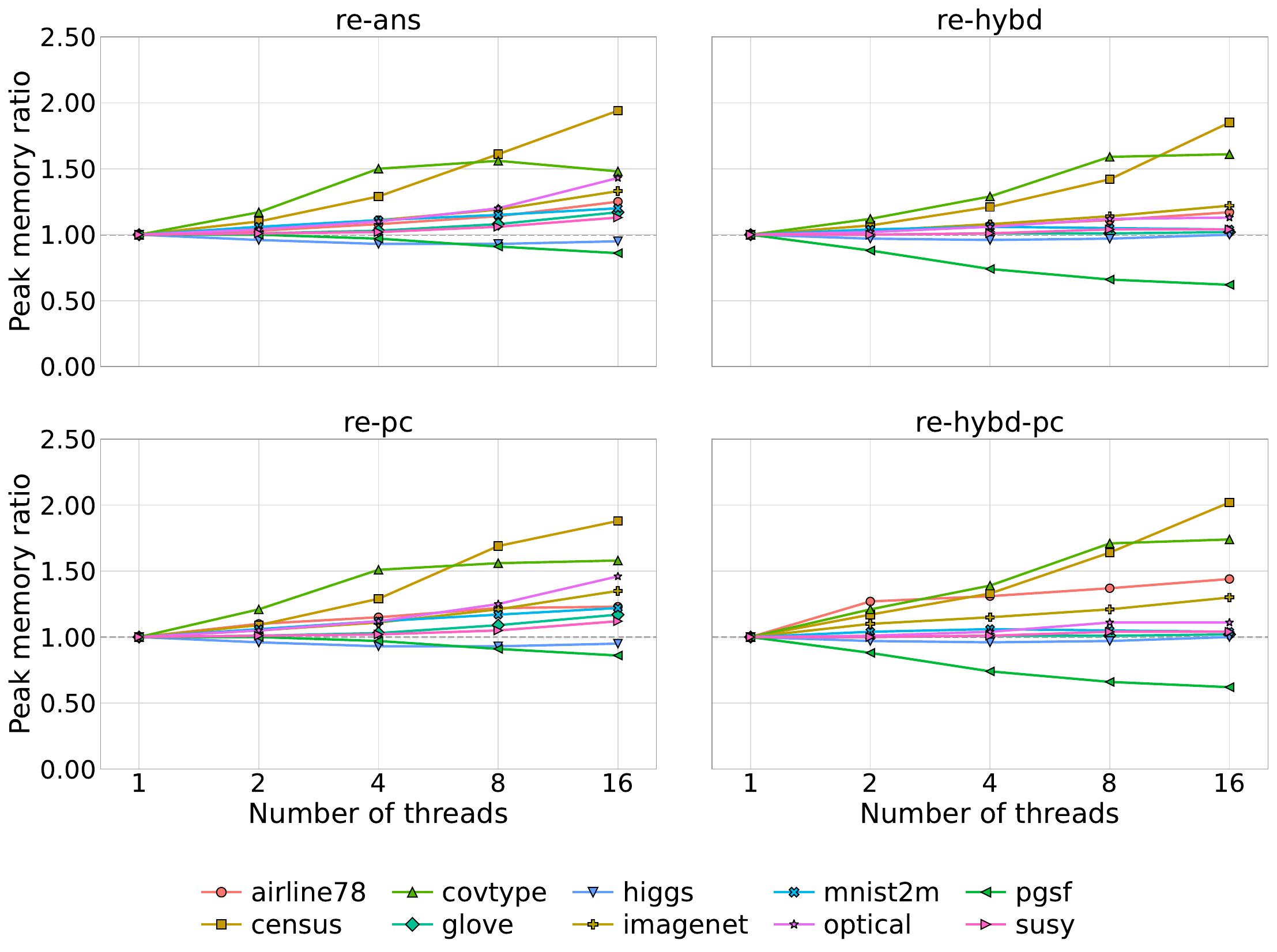}
\caption{Peak memory during matrix--vector multiplication with $p$ threads, relative to the single-threaded execution, for \reans, \reansPC, \our and \ourPC.}
\label{fig:parallel-memory}
\end{figure}

\begin{figure}[t]
\centering
\includegraphics[width=0.975\linewidth]{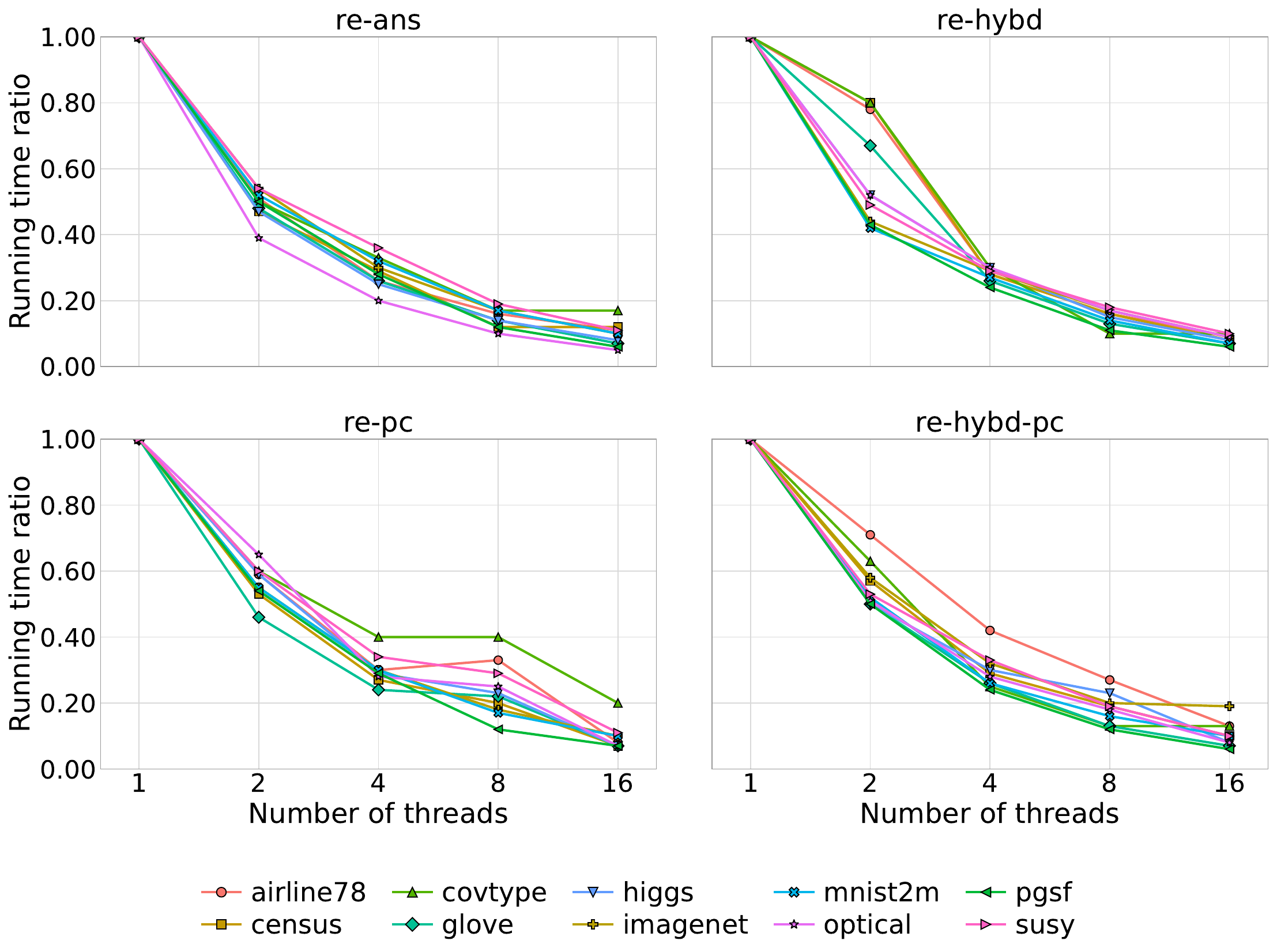}
\caption{Running time of matrix--vector multiplication with $p$ threads, relative to the single-threaded execution, for \reans, \reansPC, \our and \ourPC.}
\label{fig:parallel-time}
\end{figure}

We now evaluate the behavior of all algorithms when the input is divided into $p$ row blocks and processed with $p$ threads, as described in Section~\ref{sec:parallel}.
Figures~\ref{fig:parallel-compression}, \ref{fig:parallel-memory} and~\ref{fig:parallel-time} report the compressed size, the peak memory and the running time relative to the single-threaded execution for $p\in\{1,2,4,8,16\}$, and Table~\ref{tab:split} shows how the number of blocks changes the partition of Definition~\ref{def:split}.
Thus, the measurements combine the effects of parallel execution and changes in the compressed representation induced by the block-wise partitioning.

\paragraph{Compressed size.} Figure~\ref{fig:parallel-compression} shows that compressing the row blocks independently generally costs some compression, because repetitions spanning block boundaries are lost.
With 16 threads the compressed size of \reans grows by up to 20\% (\texttt{census}), 18\% (\texttt{covtype}) and 12\% (\texttt{imagenet}), and that of \our by at most 18\% (\texttt{census}), 8\% (\texttt{covtype}) and 5\% (\texttt{airline78} and \texttt{imagenet}).
The growth of \our never exceeds that of \reans, since part of what the grammar loses is recovered by \ansfold on the larger $S_B$.
On \pgsf the size of \our even decreases with more blocks, to 76\% of its single-thread value, from 25.5\% to 19.5\% of the dense matrix, and slightly on \texttt{higgs}, to 98\%.
Table~\ref{tab:split} explains why: digram frequencies are counted within each block, so the fraction of the sequence assigned to $S_B$ grows with $p$, dramatically so on \pgsf, from 32.3\% with one thread to 96.9\% with 16, and on these matrices the entries moved to $S_B$ are encoded more compactly by \ansfold than by the grammar.
The size of \reansPC grows like that of \reans, by up to 28\% on \texttt{covtype} and 19\% on \texttt{census}, except on \texttt{airline78}, where it drops between 8 and 16 threads, from 1.12 to 1.01 of its single-thread value, 
which is consistent with the results of~\cite{FerraginaMGKNST22}, where the reordered 16-block representation of this matrix is also smaller than the best reordering of the whole matrix.
The reordering loses its benefit with blocks in \ourPC, however: on \texttt{airline78}, \texttt{census} and \texttt{covtype} its size grows by 27\%, 39\% and 27\% at 16 threads, so that from 2 threads on it equals that of \our.
The reason is that \ourPC computes the permutation of each block on its $S_A$ alone, which shrinks as $p$ grows (Table~\ref{tab:split}), so the reordering has less to work with, whereas \reansPC reorders the whole block.

\paragraph{Peak memory.} Peak memory generally grows with $p$, since each thread keeps its own working structures, but less so for \our than for \reans: at 16 threads its ratio is at most 1.22 on eight of the ten matrices, against up to 1.43 for \reans on the same eight (Figure~\ref{fig:parallel-memory}).
On \texttt{glove}, \texttt{higgs}, \texttt{mnist2m} and \texttt{susy} it grows by at most 4\%, while that of \reans grows by 13\% to 20\% on \texttt{glove}, \texttt{mnist2m} and \texttt{susy}, since the grammar of \our, and hence the per-thread array $W$, is smaller.
The growth is pronounced for both only on \texttt{census} and \texttt{covtype}, whose single-threaded footprint is so small, about 10\% of the dense matrix or less, that the per-thread structures account for most of the total, with \our ending slightly below \reans on \texttt{census} (1.85 against 1.94) and above it on \texttt{covtype} (1.61 against 1.48).
The opposite behavior is observed on \pgsf, where the ratio of \our decreases to 0.62 at 16 threads, against 0.86 for \reans, since the shift of entries from $S_A$ to $S_B$ shrinks the grammar and $W$.
\reansPC tracks \reans, and \ourPC tracks \our except on the sparse matrices, where its ratio is higher, for the same reason as its compressed size.

\paragraph{Running time.} Both \reans and \our scale well: at 16 threads the running-time ratio lies between 0.05 and 0.17 for \reans and between 0.06 and 0.10 for \our, that is, speedups of about $6\times$ to $20\times$ and $10\times$ to $17\times$ (Figure~\ref{fig:parallel-time}).
The speedup of \our is at least as large as that of \reans on every matrix except \texttt{optical} (0.09 against 0.05), and larger on \texttt{airline78}, \texttt{census}, \texttt{covtype} and \texttt{mnist2m}, so that its single-thread slowdown with respect to \reans (Table~\ref{tab:running-time}) shrinks with $p$ on these matrices.
A likely reason is that the sparse product over $S_B$ parallelizes perfectly across row blocks, while the evaluation of $W$ in \reans is a per-thread cost that does not shrink as fast with $p$.
\reansPC and \ourPC scale similarly, with ratios between 0.07 and 0.20 and between 0.06 and 0.19 at 16 threads.
Overall, \our preserves the parallel scalability of \reans while retaining its lower peak memory footprint.

\begin{table}[t]
\centering
\caption{Fraction of the non-delimiter symbols assigned to $S_A$ and $S_B$ by the partitioning step for 1, 2, 4, 8 and 16 threads.
Fractions of $S_B$ above 50\% in bold.}
\label{tab:split}
\setlength{\tabcolsep}{3pt}
\small
\resizebox{\textwidth}{!}{%
\begin{tabular}{l|rr|rr|rr|rr|rr}
\hline
\multirow{2}{*}{Matrix}
& \multicolumn{2}{c|}{1 thread} & \multicolumn{2}{c|}{2 threads} & \multicolumn{2}{c|}{4 threads}
& \multicolumn{2}{c|}{8 threads} & \multicolumn{2}{c}{16 threads} \\
\cline{2-11}
& $S_A$ & $S_B$ & $S_A$ & $S_B$ & $S_A$ & $S_B$ & $S_A$ & $S_B$ & $S_A$ & $S_B$ \\
\hline
{\tt airline78} & 95.54\% & 4.46\% & 95.40\% & 4.60\% & 95.07\% & 4.93\% & 94.37\% & 5.63\% & 93.25\% & 6.75\% \\
{\tt census}    & 96.80\% & 3.20\% & 96.80\% & 3.20\% & 96.80\% & 3.20\% & 96.80\% & 3.20\% & 96.80\% & 3.20\% \\
{\tt covtype}   & 86.29\% & 13.71\% & 82.86\% & 17.14\% & 79.24\% & 20.76\% & 75.91\% & 24.09\% & 71.78\% & 28.22\% \\
{\tt imagenet}  & 99.55\% & 0.45\% & 99.49\% & 0.51\% & 99.39\% & 0.61\% & 99.21\% & 0.79\% & 98.92\% & 1.08\% \\
{\tt mnist2m}   & 97.81\% & 2.19\% & 95.43\% & 4.57\% & 90.30\% & 9.70\% & 82.19\% & 17.81\% & 72.97\% & 27.03\% \\
\hdashline
{\tt glove}     & 0.33\% & \textbf{99.67\%} & 0.33\% & \textbf{99.67\%} & 0.33\% & \textbf{99.67\%} & 0.33\% & \textbf{99.67\%} & 0.33\% & \textbf{99.67\%} \\
{\tt higgs}     & 27.64\% & \textbf{72.36\%} & 23.98\% & \textbf{76.02\%} & 22.54\% & \textbf{77.46\%} & 21.91\% & \textbf{78.09\%} & 21.58\% & \textbf{78.42\%} \\
{\tt optical}   & 32.26\% & \textbf{67.74\%} & 30.91\% & \textbf{69.09\%} & 28.97\% & \textbf{71.03\%} & 27.28\% & \textbf{72.72\%} & 25.11\% & \textbf{74.89\%} \\
{\pgsf}         & 67.72\% & 32.28\% & 42.81\% & \textbf{57.19\%} & 20.97\% & \textbf{79.03\%} & 8.46\% & \textbf{91.54\%} & 3.10\% & \textbf{96.90\%} \\
{\tt susy}      & 5.05\% & \textbf{94.95\%} & 5.05\% & \textbf{94.95\%} & 5.05\% & \textbf{94.95\%} & 5.05\% & \textbf{94.95\%} & 5.05\% & \textbf{94.95\%} \\
\hline
\end{tabular}}
\end{table}


\section{Related work}\label{sec:related}

Several approaches have been proposed for compressing matrices while reducing the cost of subsequent computation.
The Compressed Linear Algebra framework~\cite{Elgohary2018,Elgohary2019}, integrated in Apache SystemML, compresses column groups with lightweight encodings (offset lists, run-length and dictionary encodings) chosen by a cost model, and executes linear-algebra primitives directly on the encoded columns.
\mmRepair \cite{FerraginaMGKNST22} replaces these encodings with a single \Repair grammar over the CSRV sequence, and was shown to be more compact than CLA on the same matrices.
Its space is bounded by the $k$-th order entropy of the CSRV sequence.
The energy implications of computing over lossless compressed matrix formats are studied in~\cite{TosoniBBAFM25}.
Our work builds directly on \mmRepair and keeps its multiplication algorithm for the grammar component.
The recent survey by Navarro~\cite{Navarro26survey} classifies mm-RePair within the grammar-based family of computation-friendly matrix formats, alongside $k^2$-trees, referencing formats, and biclique decompositions.
For moderately compressible matrices under tight memory, it recommends combining such a format with a lightweight entropy coder, as \reans does for its final sequence, and \our pushes this idea further by deciding, entry by entry, which part of the matrix is left to the entropy coder.

Tosoni and Mencagli~\cite{Tosoni2026} recently ported the right product on grammar-compressed matrices to GPUs.
They make the grammar properly layered, by inserting pass-through nodes so that each level reads only the level below, and evaluate it level by level with two alternating buffers, so that the working memory depends on the width of the levels rather than on the size of the grammar.
On genotype matrices the device footprint is 3.5 to 8 times smaller than that of a CSR baseline, with running times within a small factor of \texttt{cuSPARSE}~\cite{cusparse} and lower energy, and the same engine evaluates semiring products on graphs with up to a billion edges.
Their parallelization works within a single grammar, whereas \mmRepair and \our split the matrix into row blocks, so the two approaches can be combined.
Their engine would also benefit from the smaller grammar $\mathcal{R}_A$ of \our, while the entropy-coded component $S_B$ would require a GPU decoder such as the one discussed next.

Entropy coding of sparse matrices for fast decoding has also been studied.
In~\cite{11575331} the authors propose a variant of the ANS compressor for sparse matrices that is tailored for fast parallel decoding on GPUs, and use it to achieve a significant speedup over the sparse matrix--vector multiplication routines of NVIDIA's \texttt{cuSPARSE} library~\cite{cusparse}.
Their representation is entirely entropy coded and targets bandwidth on the GPU, whereas ours combines entropy coding with grammar compression and targets the memory footprint on CPUs.
The two are complementary, and the $S_B$ component of \our could in principle be encoded with their GPU-friendly variant of the ANS compressor.

A completely different approach to matrix compression, not involving grammars or entropy coders, is the two-dimensional block tree~\cite{brisaboaTwoDimensionalBlockTrees}, which has good theoretical compression guarantees~\cite{carfagnaLandscapeCompressibilityMeasures2024} but does not support matrix--vector multiplication over the compressed data.
For the special case in which the matrix is binary, many approaches have been proposed over the years~\cite{brisaboaCompactRepresentationWeb2014,hernandezCompressedRepresentationsWeb2014,franciscoGraphCompressionAdjacencyMatrix2022a,DBLP:conf/spire/CarmonaM25,farinaCacheFriendlyCompressedBoolean2026,tabeiScalablePartialLeast2016}, including $k^2$-trees and their variants, some of them also supporting efficient matrix--vector multiplication.
These results are orthogonal to ours because we are especially interested in real-valued matrices with many distinct entries.

Lossy compression, finally, is common for the weight matrices of neural networks, which are pruned to make them sparse~\cite{HoeflerABDP21}, quantized to reduce the precision of their values~\cite{GholamiKDYMK21}, or both, as in Deep Compression~\cite{HanMD16}, whose quantized weights are then entropy coded and decoded at inference time~\cite{MarinoFMF23}.
These techniques change the matrix and may affect accuracy in a data- and model-dependent way.
Navarro~\cite{Navarro26survey} notes that combining quantization with lossless computation-friendly formats is a natural but unexplored experiment at scale, and the quantized variants of Section~\ref{sec:quantization} are a first step in this direction, which shows that the gain of \our over \mmRepair holds at every quantization level and is largest at intermediate ones.



\section{Conclusion and future work}\label{sec:conclusion}

We presented \our, a hybrid strategy for the \mmRepair framework that combines grammar and entropy compression, yielding more compact representations that still support matrix--vector multiplication directly over the compressed data.
A simple partitioning rule, justified by an accounting of when a grammar rule actually saves space, routes to an entropy coder the entries that \Repair would leave as terminals.
Row ordering with delta encoding and a cost-driven alphabet remapping make the entropy-coded parts smaller still.
Experiments on ten machine-learning matrices showed that \our improves compression ratios and peak memory over \mmRepair on every input, most notably on less sparse matrices and matrices with many distinct values, while keeping multiplication times competitive.
Column reordering, in the \reansPC variant of \mmRepair, is complementary to the partitioning: \our outperforms \reansPC on most of the matrices, and on the sparse matrices their combination, \ourPC, is even more compact, at the price of a much longer compression time.
Finally, \our scales with the number of threads as well as \mmRepair, while retaining its lower memory footprint.

Several directions remain open.
A cheaper encoding of $S_B$ could reduce the running-time overhead of \our on some inputs, at the cost of a little space.
The value array $V$ is stored uncompressed and takes a significant part of the representation on matrices with many distinct values, so compressing it would help the inputs on which \our is most effective.
The grammar component could be built by other grammar compressors (\eg~\cite{BilleGP17,GagieIMNST19,TakabatakeIS17,NunesLGAN22,AngeloNTL26}), provided that their rules do not cross row delimiters and that the multiplication algorithm is adapted to rules of arbitrary length.
Finally, a GPU implementation could combine the level-by-level grammar evaluation of~\cite{Tosoni2026} for $\mathcal{R}_A$ with a parallel entropy decoder for $S_B$, such as that of~\cite{11575331}, evaluating the two components concurrently and summing their results.


\section*{Acknowledgments}
F.A.L.\ and G.P.T.\ were partially supported by the Brazilian National Council for
Scientifici and Technological Development (CNPq), grants 408314/2023-0, 313070/2025-3, 311128/2025-4 and 351599/2025-8, and by the São Paulo Research Foundation (FAPESP), process 2025/16208-7.
G.M.\ was partially supported by INdAM-GNCS Project CUP E53C25002010001 and PAN-HUB Project, POS 2014--2020, ID T4-AN-07, CUP I53C22001300001.

\section*{Declaration of competing interest}
The authors declare that they have no known competing financial interests or personal relationships that could have appeared to influence the work reported in this paper.

\section*{Declaration of generative AI and AI-assisted technologies in the writing process}
During the preparation of this work the authors used Claude (Anthropic) in order to improve the readability and language of the manuscript and to assist in the preparation of figures.
After using this tool, the authors reviewed and edited the content as needed and take full responsibility for the content of the publication.

\section*{Data availability}
The code is available at \url{https://github.com/felipelouza/mm-repair-hybrid/} and the datasets at \url{https://www.kaggle.com/datasets/giovannimanzini/some-machine-learning-matrices}.

\bibliographystyle{plain}
\bibliography{refs}

@string{ spire  = "Proc. International Symposium on String Processing and Information Retrieval (SPIRE)" }

@string{ alenex  = "Proc. Workshop on Algorithm Engineering and Experimentation (ALENEX)" }

@string{ dcc   = "Proc. IEEE Data Compression Conference (DCC)" }

@string{ esa = "Proc. European Symposium on Algorithms (ESA)" }

@string{ sea = "Proc. SEA" }

@string{ vldb = "Proc. International Conference on Very Large Data Bases (VLDB)" }

@inproceedings{Gog2014a,
	author = {Gog, Simon and Beller, Timo and Moffat, Alistair and Petri, Matthias},
	booktitle = sea,
	pages = {326--337},
	title = {From Theory to Practice: Plug and Play with Succinct Data Structures},
	year = {2014}
}

@book{Navarro2016,
	author = {Gonzalo Navarro},
	title = {Compact Data Structures -- A practical approach},
	publisher = {Cambridge University Press},
	year = 2016
}

@article{NunesLGAN22,
  author       = {Daniel Saad Nogueira Nunes and
                  Felipe A. Louza and
                  Simon Gog and
                  Mauricio Ayala{-}Rinc{\'{o}}n and
                  Gonzalo Navarro},
  title        = {Grammar Compression by Induced Suffix Sorting},
  journal      = {{ACM} J. Exp. Algorithmics},
  volume       = {27},
  pages        = {1.1:1--1.1:33},
  year         = {2022}
}

@article{CharikarLLPPSS05,
  author       = {Moses Charikar and
                  Eric Lehman and
                  Ding Liu and
                  Rina Panigrahy and
                  Manoj Prabhakaran and
                  Amit Sahai and
                  Abhi Shelat},
  title        = {The smallest grammar problem},
  journal      = {{IEEE} Trans. Inf. Theory},
  volume       = {51},
  number       = {7},
  pages        = {2554--2576},
  year         = {2005}
}

@ARTICLE{repair,
  author       = {N. Jesper Larsson and
                  Alistair Moffat},
  title        = {Off-line dictionary-based compression},
  journal      = {Proc. {IEEE}},
  volume       = {88},
  number       = {11},
  pages        = {1722--1732},
  year         = {2000}
}

@article{FerraginaMGKNST22,
  author       = {Paolo Ferragina and
                  Giovanni Manzini and
                  Travis Gagie and
                  Dominik K{\"{o}}ppl and
                  Gonzalo Navarro and
                  Manuel Striani and
                  Francesco Tosoni},
  title        = {Improving Matrix-vector Multiplication via Lossless Grammar-Compressed
                  Matrices},
  journal      = {Proc. {VLDB} Endow.},
  volume       = {15},
  number       = {10},
  pages        = {2175--2187},
  year         = {2022},
}

@inproceedings{GagieIMNST19,
  author       = {Travis Gagie and
                  Tomohiro I and
                  Giovanni Manzini and
                  Gonzalo Navarro and
                  Hiroshi Sakamoto and
                  Yoshimasa Takabatake},
  title        = {Rpair: Rescaling RePair with Rsync},
  booktitle    = spire,
  series       = {LNCS},
  volume       = {11811},
  pages        = {35--44},
  publisher    = {Springer},
  year         = {2019},
  doi          = {10.1007/978-3-030-32686-9\_3},
}

@article{Elgohary2019,
  title={Compressed Linear Algebra for Declarative Large-Scale Machine Learning},
  author={Elgohary, Ahmed and Boehm, Matthias and Haas, Peter J. and Reiss, Frederick R. and Reinwald, Berthold},
  journal={Communications of the ACM},
  volume={62},
  number={5},
  pages={83--91},
  year={2019}
}

@article{Elgohary2018,
  title={Compressed linear algebra for large-scale machine learning},
  author={Elgohary, Ahmed and Boehm, Matthias and Haas, Peter J. and Reiss, Frederick R. and Reinwald, Berthold},
  journal={The VLDB Journal},
  volume={27},
  number={5},
  pages={719--744},
  year={2018}
}

@article{Larsson2000,
  title={Off-line dictionary-based compression},
  author={Larsson, N. Jesper and Moffat, Alistair},
  journal={Proceedings of the IEEE},
  volume={88},
  number={11},
  pages={1722--1732},
  year={2000}
}

@article{Moffat2020,
  title={Large-Alphabet Semi-Static Entropy Coding Via Asymmetric Numeral Systems},
  author={Moffat, Alistair and Petri, Matthias},
  journal={{ACM} Trans. Inf. Syst.},
  volume={38},
  number={4},
  pages={33:1--33:33},
  year={2020}
}

@article{TosoniBBAFM25,
  author       = {Francesco Tosoni and
                  Philip Bille and
                  Valerio Brunacci and
                  Alessio De Angelis and
                  Paolo Ferragina and
                  Giovanni Manzini},
  title        = {Toward Greener Matrix Operations by Lossless Compressed Formats},
  journal      = {{IEEE} Access},
  volume       = {13},
  pages        = {56756--56779},
  year         = {2025}
}

@article{Lohrey12,
  author       = {Markus Lohrey},
  title        = {Algorithmics on {SLP}-compressed strings: {A} survey},
  journal      = {Groups Complex. Cryptol.},
  volume       = {4},
  number       = {2},
  pages        = {241--299},
  year         = {2012}
}

@inproceedings{pennington-etal-2014-glove,
    title     = "{G}lo{V}e: Global Vectors for Word Representation",
    author    = "Pennington, Jeffrey and Socher, Richard and Manning, Christopher",
    booktitle = "Proc. EMNLP",
    month     = oct,
    year      = "2014",
    address   = "Doha, Qatar",
    publisher = "Association for Computational Linguistics",
    pages     = "1532--1543",
}

@article{Wang2020,
  title = {A Survey on Large-Scale Machine Learning},
  ISSN = {2326-3865},
  DOI = {10.1109/tkde.2020.3015777},
  journal = {IEEE Transactions on Knowledge and Data Engineering},
  publisher = {Institute of Electrical and Electronics Engineers (IEEE)},
  author = {Wang,  Meng and Fu,  Weijie and He,  Xiangnan and Hao,  Shijie and Wu,  Xindong},
  year = {2020},
  pages = {1–1}
}

@INPROCEEDINGS {11575331,
author = { Schatzle, Emil and Pegolotti, Tommaso and Puschel, Markus },
booktitle = {Proc. IPDPS},
title = {{ Fast Entropy Decoding for Sparse MVM on GPUs }},
year = {2026},
volume = {},
ISSN = {},
pages = {416-432},
doi = {10.1109/IPDPS65963.2026.00044},
publisher = {IEEE Computer Society}
}

@article{hernandezCompressedRepresentationsWeb2014,
    title = {Compressed representations for web and social graphs},
    volume = {40},
    issn = {0219-3116},
    doi = {10.1007/s10115-013-0648-4},
    number = {2},
    urldate = {2024-07-25},
    journal = {Knowledge and Information Systems},
    author = {Hernández, Cecilia and Navarro, Gonzalo},
    month = aug,
    year = {2014},
    pages = {279--313},
}

@article{franciscoGraphCompressionAdjacencyMatrix2022a,
    title = {Graph {Compression} for {Adjacency}-{Matrix} {Multiplication}},
    volume = {3},
    issn = {2662-995X, 2661-8907},
    doi = {10.1007/s42979-022-01084-2},
    number = {3},
    urldate = {2026-08-17},
    journal = {{SN} Comput. Sci.},
    author = {Francisco, Alexandre P. and Gagie, Travis and Köppl, Dominik and Ladra, Susana and Navarro, Gonzalo},
    year = {2022},
    pages = {193},
}

@inproceedings{tabeiScalablePartialLeast2016,
    address = {New York, NY, USA},
    title = {Scalable {Partial} {Least} {Squares} {Regression} on {Grammar}-{Compressed} {Data} {Matrices}},
    isbn = {978-1-4503-4232-2},
    doi = {10.1145/2939672.2939864},
    booktitle = {Proc. SIGKDD},
    publisher = {Association for Computing Machinery},
    author = {Tabei, Yasuo and Saigo, Hiroto and Yamanishi, Yoshihiro and Puglisi, Simon J.},
    month = aug,
    year = {2016},
    pages = {1875--1884},
}

@inproceedings{DBLP:conf/spire/CarmonaM25,
  author       = {Gabriel Carmona and
                  Giovanni Manzini},
  title        = {Depth First Representations of $k^2$-trees},
  booktitle    = {Proc. SPIRE},
  series       = {LNCS},
  volume       = {16073},
  pages        = {28--44},
  publisher    = {Springer},
  year         = {2025},
  doi          = {10.1007/978-3-032-05228-5\_4}
}

@article{brisaboaTwoDimensionalBlockTrees,
  title = {Two-{{Dimensional Block Trees}}},
  author = {Brisaboa, Nieves R and Gagie, Travis and {Gomez-Brandon}, Adrian and Navarro, Gonzalo},
  year = 2024,
  journal = {The Computer Journal},
  volume = {67},
  langid = {english},
}

@inproceedings{yu2023compressing,
  title     = {Compressing Transformers: Features Are Low-Rank, but Weights Are Not!},
  author    = {Yu, Hao and Wu, Jianxin},
  booktitle = {Proc. AAAI},
  volume    = {37},
  number    = {9},
  pages     = {11007--11015},
  year      = {2023},
  doi       = {10.1609/aaai.v37i9.26304}
}

@inproceedings{LouzaMT2027,
        author = {Louza, Felipe Alves and Manzini, Giovanni and Telles, Guilherme Pimentel},
        booktitle = dcc,
        title = {Better Matrix–Vector Multiplication via
Hybrid Matrix Compression},
        note          = {To appear},
        year = {2027}
}

@article{KiefferY00,
  author       = {John C. Kieffer and
                  En{-}Hui Yang},
  title        = {Grammar-based codes: {A} new class of universal lossless source codes},
  journal      = {{IEEE} Trans. Inf. Theory},
  volume       = {46},
  number       = {3},
  pages        = {737--754},
  year         = {2000},
  url          = {https://doi.org/10.1109/18.841160},
  doi          = {10.1109/18.841160},
  bibsource    = {dblp computer science bibliography, https://dblp.org}
}

@article{OchoaN19,
  author       = {Carlos Ochoa and
                  Gonzalo Navarro},
  title        = {RePair and All Irreducible Grammars are Upper Bounded by High-Order
                  Empirical Entropy},
  journal      = {{IEEE} Trans. Inf. Theory},
  volume       = {65},
  number       = {5},
  pages        = {3160--3164},
  year         = {2019},
  url          = {https://doi.org/10.1109/TIT.2018.2871452},
  doi          = {10.1109/TIT.2018.2871452},
  bibsource    = {dblp computer science bibliography, https://dblp.org}
}

@article{Navarro26survey,
  author  = {Gonzalo Navarro},
  title   = {Computation-Friendly Compression of Matrices: Speeding up Linear and {Boolean} Algebra},
  journal = {arXiv preprint arXiv:2610.35362},
  year    = {2026}
}

@book{Saad2003,
  author       = {Yousef Saad},
  title        = {Iterative methods for sparse linear systems},
  publisher    = {{SIAM}},
  year         = {2003},
  doi          = {10.1137/1.9780898718003}
}

@inproceedings{KourtisGK08,
  author       = {Kornilios Kourtis and
                  Georgios I. Goumas and
                  Nectarios Koziris},
  editor       = {Alex Ram{\'{\i}}rez and
                  Gianfranco Bilardi and
                  Michael Gschwind},
  title        = {Optimizing sparse matrix-vector multiplication using index and value
                  compression},
  booktitle    = {Proceedings of the 5th Conference on Computing Frontiers, 2008, Ischia,
                  Italy, May 5-7, 2008},
  pages        = {87--96},
  publisher    = {{ACM}},
  year         = {2008},
  doi          = {10.1145/1366230.1366244}
}

@inproceedings{Tosoni2026,
 title         = {Streaming Right Multiplication over Grammar-Compressed Matrices: A Memory-Bounded {GPU} Engine for Genotype and Graph Data},
 author        = {Tosoni, Francesco and Mencagli, Gabriele},
 booktitle     = {Proceedings of the SIAM Symposium on Algorithm Engineering and Experiments (ALENEX)},
 year          = {2027},
 publisher     = {SIAM},
 note          = {To appear},
 eprint        = {2607.24971},
 archivePrefix = {arXiv},
 primaryClass  = {cs.MS}
}

@article{HanMD16,
  author  = {Song Han and Huizi Mao and William J. Dally},
  title   = {Deep Compression: Compressing Deep Neural Networks with Pruning, Trained Quantization and {H}uffman Coding},
  journal = {Proc. International Conference on Learning Representations (ICLR)},
  year    = {2016}
}

@misc{GholamiKDYMK21,
  author       = {Amir Gholami and
                  Sehoon Kim and
                  Zhen Dong and
                  Zhewei Yao and
                  Michael W. Mahoney and
                  Kurt Keutzer},
  title        = {A Survey of Quantization Methods for Efficient Neural Network Inference},
  journal      = {CoRR},
  volume       = {abs/2103.13630},
  year         = {2021},
  doi          = {10.48550/ARXIV.2103.13630},
  eprinttype   = {arXiv},
  eprint       = {2103.13630}
}

@article{HoeflerABDP21,
  author       = {Torsten Hoefler and
                  Dan Alistarh and
                  Tal Ben{-}Nun and
                  Nikoli Dryden and
                  Alexandra Peste},
  title        = {Sparsity in Deep Learning: Pruning and growth for efficient inference
                  and training in neural networks},
  journal      = {J. Mach. Learn. Res.},
  volume       = {22},
  pages        = {241:1--241:124},
  year         = {2021}
}

@article{MarinoFMF23,
  author       = {Giosu{\`{e}} Cataldo Marin{\`{o}} and
                  Flavio Furia and
                  Dario Malchiodi and
                  Marco Frasca},
  title        = {Efficient and Compact Representations of Deep Neural Networks via
                  Entropy Coding},
  journal      = {{IEEE} Access},
  volume       = {11},
  pages        = {106103--106125},
  year         = {2023},
  doi          = {10.1109/ACCESS.2023.3317293}
}

@article{AngeloNTL26,
  author       = {Danyelle S. O. Angelo and
                  Daniel Saad Nogueira Nunes and
                  Guilherme P. Telles and
                  Felipe A. Louza},
  title        = {Grammar Compression With Support to Fast Substring Extraction},
  journal      = {{IEEE} Access},
  volume       = {14},
  pages        = {36440--36452},
  year         = {2026}
}

@inproceedings{TakabatakeIS17,
  author    = {Yoshimasa Takabatake and Tomohiro I and Hiroshi Sakamoto},
  title     = {A Space-Optimal Grammar Compression},
  booktitle = {Proc. 25th Annual European Symposium on Algorithms (ESA)},
  series    = {LIPIcs},
  volume    = {87},
  pages     = {67:1--67:15},
  year      = {2017}
}

@inproceedings{BilleGP17,
  author    = {Philip Bille and Inge Li G{\o}rtz and Nicola Prezza},
  title     = {Space-Efficient {Re-Pair} Compression},
  booktitle = {Proc. Data Compression Conference (DCC)},
  pages     = {171--180},
  year      = {2017}
}

@misc{cusparse,
  author       = {{NVIDIA Corporation}},
  title        = {{cuSPARSE} Library},
  howpublished = {\url{https://developer.nvidia.com/cusparse}},
  note         = {Accessed: 2026-11-12}
}

@incollection{farinaCacheFriendlyCompressedBoolean2026,
    address = {Cham},
    title = {Cache-{Friendly} {Compressed} {Boolean} {Matrices}},
    volume = {16073},
    isbn = {978-3-032-05227-8 978-3-032-05228-5},
    doi = {10.1007/978-3-032-05228-5\_9},
    language = {en},
    booktitle = {Proc. SPIRE},
    publisher = {Springer Nature Switzerland},
    author = {Fariña, Antonio and Gómez-Brandón, Adrián and Gómez-Colomer, Asunción and Navarro, Gonzalo},
    year = {2026},
    note = {LNCS},
    pages = {95--108},
}

@article{brisaboaCompactRepresentationWeb2014,
    title = {Compact representation of {Web} graphs with extended functionality},
    volume = {39},
    issn = {03064379},
    doi = {10.1016/j.is.2013.08.003},
    language = {en},
    urldate = {2024-04-13},
    journal = {Information Systems},
    author = {Brisaboa, Nieves R. and Ladra, Susana and Navarro, Gonzalo},
    month = jan,
    year = {2014},
    pages = {152--174},
}

@article{carfagnaLandscapeCompressibilityMeasures2024,
    title = {The {Landscape} of {Compressibility} {Measures} for {Two}-{Dimensional} {Data}},
    volume = {12},
    copyright = {https://creativecommons.org/licenses/by-nc-nd/4.0/},
    issn = {2169-3536},
    doi = {10.1109/ACCESS.2024.3417621},
    language = {en},
    urldate = {2026-06-02},
    journal = {IEEE Access},
    author = {Carfagna, Lorenzo and Manzini, Giovanni},
    year = {2024},
    pages = {87268--87283},
}

\end{document}